\documentclass[]{prlab}

\usepackage[toc,page,header]{appendix}
\microtypesetup{expansion=false}
\usepackage{graphicx}
\usepackage{subcaption}
\usepackage{booktabs}
\usepackage{hyperref}
\usepackage{bm}
\usepackage{amsmath}
\usepackage{amssymb}
\usepackage{mathtools}
\usepackage{amsthm}
\usepackage{xspace}
\usepackage{wrapfig}
\usepackage{multirow}
\usepackage{tabularx}
\usepackage{colortbl}
\usepackage{pifont}
\usepackage{xcolor}
\usepackage{eso-pic}
\definecolor{deepgreen}{RGB}{0,100,0}

\title{{\color{seedblue}\slshape AURORA-LM}: Autoencoding Unified Representation for Continuous-Latent Diffusion Language Modeling}

\author[1,\star]{Jiajun Liang}
\author[1,\star]{Yucheng Liao}
\author[2,\star]{Yukang Cao}
\author[1]{Jiazhe Wei}
\author[1]{Ken Li}
\author[3]{Wende Tan}
\author[1]{Jiankun Zhang}
\author[1]{ZY Cui}
\author[1]{Jingkang~Yang}
\author[3]{Liucheng Guo}
\author[1]{Shiqi Yang}
\author[]{B. Yang}
\author[1]{Caifeng Shan}
\author[2]{Ziwei Liu}
\author[1,\dagger]{Chenyang~Si}

\affiliation[1]{PRLab, Nanjing University}
\affiliation[2]{S-Lab, Nanyang Technological University}
\affiliation[3]{Imperial College London}

\contribution[\star]{Equal Contribution}
\contribution[\dagger]{Corresponding Author}

\renewcommand{\affiliationfont}{\fontsize{9.5}{11}\selectfont}

\checkdata[Code]{\url{https://github.com/fyv587/AURORA-LM}}
\checkdata[Project Page]{\url{https://aurora-lm-project.github.io/}}

\abstract{
Language remains an outlier in modern generative modeling: while images, video, and audio are increasingly modeled in continuous latent spaces, text generation still relies predominantly on discrete tokens. Existing continuous language models either inherit embedding spaces not designed for generation and decoding jointly, or compress autoencoded latents to make diffusion easier at the cost of token-level fidelity.
We challenge this prevailing design compromise.
Instead of simplifying the representation to accommodate the generative model, we preserve a high-capacity, decodable text latent and design the diffusion model to learn its distribution directly.\par\vspace{0.6em}

We introduce \textbf{AURORA-LM}, a continuous-latent diffusion language model that separates the construction of a decodable text representation from the modeling of its generative distribution.

To obtain such a representation, we use a Query-based Encoder-Decoder that organizes text into a high-capacity, prefix-aligned latent sequence.
We then introduce a Block-causal Diffusion Transformer that learns the distribution of these full-width latents through flow matching, generating blocks from left to right while denoising the positions within each block in parallel.
However, retaining a high-capacity latent representation for accurate token decoding also makes its distribution more challenging for the diffusion model to learn.
AURORA-LM addresses this difficulty by restricting only the noisy-input pathway while retaining the full clean-latent prediction target, allowing the generative model to accommodate the full-width latent without reducing decoder-facing capacity.
We further calibrate the noise-level distribution to the latent width, accounting for how the effective signal strength changes with representation dimensionality. 
Finally, we introduce self-trajectory consistency to bridge the gap between training on independently sampled noisy states and inference through iterative denoising.\par\vspace{0.6em}

Across comprehensive comparisons, AURORA-LM achieves the strongest performance among the evaluated continuous and diffusion-based language models on OpenWebText free generation and XSum conditional summarization. 
Scaling to 1B parameters with approximately 1500 EFLOPs of total compute yields further gains and surpasses a larger publicly released latent-diffusion language model under a matched evaluation protocol. Our results demonstrate that continuous language generation can effectively bridge diffusion-based generative modeling and discrete token decoding through a high-capacity, causally structured, and decodable text representation. All experiments are conducted on Ascend NPUs.

}

\begin{document}
\AddToShipoutPictureBG*{%
  \AtTextLowerLeft{%
    \raisebox{-2.2\baselineskip}[0pt][0pt]{%
      \parbox[t]{\textwidth}{\footnotesize
        \rule{0.3\textwidth}{0.4pt}\\[2pt]
        Contact: \href{grealish2821@gmail.com}{grealish2821@gmail.com}, 
        \href{cheng_2528@Outlook.com}{cheng\_2528@Outlook.com}, 
        \href{yukang.cao@ntu.edu.sg}{yukang.cao@ntu.edu.sg}, \href{mailto:chenyang.si@nju.edu.cn}{chenyang.si@nju.edu.cn}, 
        
        }}}}
\maketitle
\newpage

\tableofcontents
\clearpage

\section{Introduction}\label{sec:intro}

Continuous latent spaces have become a general interface for representation learning and generation~\citep{rombach2022ldm}. 
Among them, diffusion model is a representative example: by learning generative dynamics in continuous spaces, they have achieved remarkable quality in image, video, and audio synthesis~\citep{ho2020ddpm,song2021scoresde,rombach2022ldm,ho2022videodiffusion,kong2020diffwave}. 
These advantages suggest that continuous spaces may serve as a general interface for generation across modalities, raising the question of whether language itself can be represented and generated within such a space.
Language, however, remains largely outside this paradigm.
Visual and acoustic signals are naturally represented in continuous spaces, whereas modern language models rely on discrete token representations~\citep{brown2020gpt3,touvron2023llama,yang2024qwen3}.

This asymmetry also shapes the architecture of current multimodal systems. Continuous observations are commonly converted into discrete token sequences before they are consumed by a 
token-based Transformer backbone~\citep{chameleon2024chameleon,wang2024emu3}.
Such conversion is useful because it reuses the infrastructure of language modeling, but it makes discreteness the universal interface and may suppress structure that is more naturally represented continuously.
The converse direction is therefore equally important: \textbf{instead of forcing every modality into discrete tokens, can language itself be represented and generated in a continuous latent space?}
A satisfactory answer would provide a direct bridge between language and continuous generative models, while retaining the semantic fidelity and syntactic regularity expected of texts.

Early attempts at continuous language generation largely begin from representations that already exist.
Diffusion-LM~\citep{li2022diffusionlm} applies diffusion directly to word
embeddings, while PLAID~\citep{gulrajani2023plaid} develops this token-wise
embedding formulation for likelihood-based language modeling.
TEncDM~\citep{shabalin2024tencdm} replaces static word vectors with contextual features extracted by a pre-trained language encoder and trains a separate decoder to reconstruct text.
These approaches establish that language can be generated through continuous dynamics rather than exclusively through categorical next-token prediction.
However, because these methods inherit their representations from token embeddings or pre-trained encoders, properties such as latent width, dimensionality, and sequence organization are not explicitly designed to support both accurate text reconstruction and effective continuous generation.

To gain direct control over the representation itself, language autoencoders offer a more flexible alternative by making the generated representation itself learnable.
Methods such as LD4LG~\citep{lovelace2023ld4lg}, COSMOS~\citep{meshchaninov2025cosmos}, and Cola-DLM~\citep{guo2026cola} first map text into a continuous latent space and then learn a diffusion- or flow-based model over that space.
This formulation allows the latent sequence to be compressed, smoothed, or organized hierarchically, making its distribution more tractable for generation.
Yet, these benefits are typically obtained by simplifying the same representation on which the decoder relies. 
A compact and smooth latent can be easier for the generative model to learn, but compression will also remove distinctions needed to recover the exact words, syntax, and local ordering.
Conversely, retaining more information improves reconstruction but presents the latent prior with a wider and more complex distribution. 
The difficulty of latent generation and the fidelity of text reconstruction therefore remain coupled through a single representational bottleneck.

This tension suggests that a continuous language representation should not be designed solely to make its distribution easy to model.
It must first serve as an adequate interface to text, faithfully preserving linguistic semantics while retaining sufficient information for accurate token recovery.
The latent generative model should then be designed to accommodate this representation, without compressing the decoder-facing latent merely to make generation easier.
Therefore, efficient continuous language generation presents two connected challenges:
constructing a continuous text representation that supports faithful and accurate decoding, and designing a generative model that can effectively learn the resulting latent distribution without sacrificing decoding fidelity.

To address these two connected challenges, we introduce \textbf{AURORA-LM} (Autoencoding Unified Representation for Continuous-Latent Diffusion Language Modeling), a unified framework that decouples representation construction and distribution modeling into two stages.
For the first challenge, we construct a high-capacity continuous latent sequence from discrete text using a Query-based Encoder-Decoder.
Its causal encoder allows successive latent positions to read progressively longer token prefixes, inducing a prefix order that supports blockwise left-to-right generation.
A corresponding decoder maps latent prefixes to token logits, forming an explicit latent-to-token pathway trained solely through text reconstruction.
Because the latent sequence serves directly as the decoder input, we retain its channel capacity for accurate token recovery rather than reducing it merely to simplify diffusion-based generation.

Having established a decodable and causally ordered latent space, we turn to the second challenge: learning its induced distribution without sacrificing the representational capacity.
Specifically, we freeze the autoencoder and train a block-causal denoiser on the resulting full-width latent sequences using flow matching, following their causal organization across positions.
The model generates latent blocks from left to right while jointly denoising positions within each block.
To effectively model this full-width latent distribution, we combine three complementary designs. 
First, inspired by JiT's direct clean-data prediction principle~\citep{li2025jit}, we apply a low-rank projection only to the noisy-latent input pathway while continuing to predict the full-width clean latent.
Second, we calibrate the training-time noise allocation to latent width. Empirically, wider latent representations benefit from placing greater emphasis on high-noise inputs.
Third, we introduce self-trajectory consistency, which aligns clean-latent predictions at neighboring states along the denoising trajectory and enables high-quality generation with fewer sampling steps.

\begin{figure}[!t]
    \centering
    \begin{minipage}[t]{0.73\textwidth}
        \vspace{0pt}
        \centering
        \includegraphics[width=\linewidth]{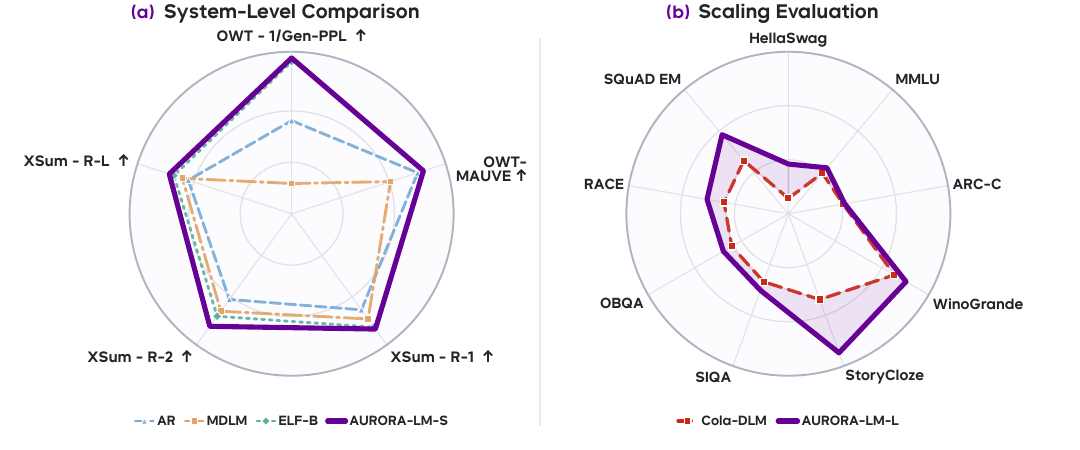}
    \end{minipage}
    \hfill
    \begin{minipage}[t]{0.25\textwidth}
        \vspace{0pt}\captionsetup{justification=raggedright,singlelinecheck=false,
            skip=0pt
        }\caption{\textbf{Evaluations of AURORA-LM across various generation settings and model scales.}
        (a) AURORA-LM achieves the best displayed results on OpenWebText and XSum; radial values are normalized per metric, with Gen-PPL shown reciprocally. ELF-B's MAUVE is omitted because it uses a different evaluation protocol.
        (b) AURORA-LM outperforms a larger publicly released latent-diffusion language model across nine language benchmarks.}
        \label{fig:teaser}
    \end{minipage}
\end{figure}

Our experiments systematically analyze the design choices that govern continuous language generation,
including latent width and sequence compression, the low-rank input bottleneck, noise-level
allocation, the prediction target and loss space, self-trajectory consistency, and block-causal
generation.
We also compare AURORA-LM with existing continuous and diffusion-based language models
under matched data and evaluation protocols.
As summarized in Fig.~\ref{fig:teaser}, AURORA-LM achieves the
strongest performance among the evaluated methods on both OpenWebText~\citep{gokaslan2019openwebtext} free generation and XSum~\citep{narayan2018xsum}
conditional summarization. 
We further scale AURORA-LM to 1B parameters with approximately 1,500 EFLOPs of
total compute, and observe consistent performance gains over a larger publicly
released latent-diffusion language model under a matched benchmark protocol.
Together, these results demonstrate that \textbf{a high-capacity, decodable text representation can provide an effective interface between continuous generation and discrete token decoding when the diffusion model is designed to learn its distribution directly.}

In summary, our contributions are as follows:
\begin{itemize}
    \item We introduce \textbf{AURORA-LM}, a unified and continuous-latent diffusion language model that separates
    text representation learning from generative distribution modeling. Instead of inheriting the
    denoising space from token embeddings or pretrained encoders, AURORA-LM explicitly constructs continuous representation that the diffusion model learns to generate.

    \item We construct a \textbf{high-capacity autoencoded text latent} that serves as the decoder-facing interface between continuous generation and token recovery. Rather than reducing this representation solely for prior tractability, AURORA-LM retains the channel capacity available for accurate decoding and adapts the diffusion prior to model the resulting full-width latent distribution.
    
    \item We learn the resulting latent distribution with a \textbf{block-causal denoiser} trained by flow matching, which generates latent blocks from left to right while jointly denoising positions within each block. To model the full-width latent target, we combine a
    low-rank noisy-input pathway with full-width clean-latent prediction, calibrate the training-time
    noise allocation to latent width, and introduce self-trajectory consistency to improve
    generation along the denoising trajectory.

    \item Under matched data and evaluation protocols, AURORA-LM outperforms the evaluated
    autoregressive, discrete-diffusion, and continuous embedding-based baselines on OpenWebText free generation and XSum conditional summarization. We also scale AURORA-LM to 1B
    parameters, where it continues to perform strongly and surpasses a larger publicly released latent-diffusion language model under the same benchmark protocol.
\end{itemize}

\section{Related Work}\label{sec:related}

\subsection{Diffusion Models}\label{sec:rel-diffusion}

Diffusion models~\citep{ho2020ddpm,song2021scoresde,karras2022edm} generate
samples by learning to invert a forward process that progressively injects
Gaussian noise into clean data. Flow matching~\citep{lipman2023flow,liu2023rectified,albergo2023stochastic}
recasts generation as learning a velocity field that continuously transports
a noise distribution to the data distribution, enabling deterministic
generation via ODE integration. Consistency models~\citep{song2023consistency,luo2023latentconsistency}
address multi-step inference cost by enforcing that all states along the same
probability-flow ODE trajectory map to a consistent clean endpoint, enabling
one- or few-step generation. AURORA-LM uses linear-interpolant flow matching
to train a latent generative model and incorporates self-trajectory consistency as a training regularizer.

\begin{figure*}[h]
    \centering
    \includegraphics[width=\textwidth]{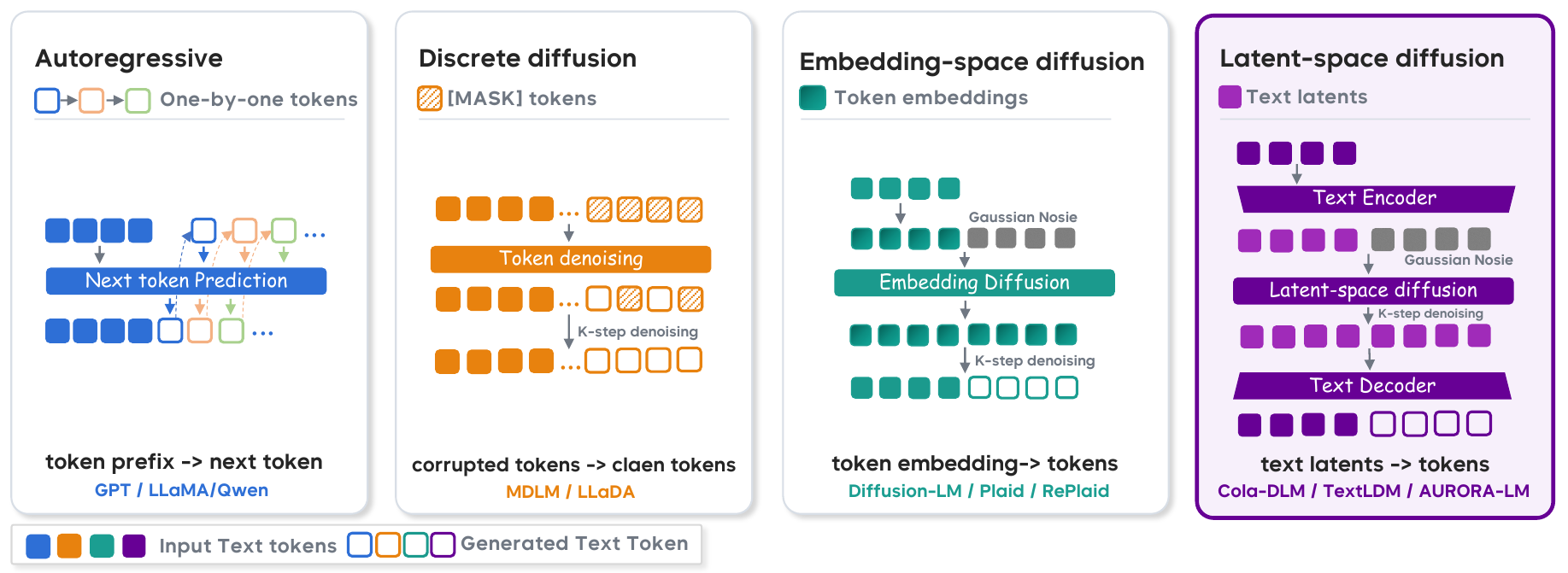}
    \caption{
    \textbf{Language modeling paradigms organized by the representation generated at inference.} Autoregressive models generate discrete tokens one at a time; masked discrete diffusion iteratively denoises masked tokens; embedding-space diffusion denoises continuous token embeddings; and latent-space diffusion generates encoder--decoder text latents before decoding them into tokens. AURORA-LM adopts the latent-space formulation.}
    \label{fig:related-paradigms}
\end{figure*}

Figure~\ref{fig:related-paradigms} surveys the resulting landscape of language
generation paradigms. We discuss each in the following subsections, from
autoregressive and discrete diffusion models to continuous latent diffusion
approaches.

\subsection{Discrete Language Models}\label{sec:rel-discrete}

\noindent\textbf{Autoregressive Language Models.}
Autoregressive language models factorize the joint distribution over a token
sequence as $p(x) = \prod_{t=1}^{L} p(x_t \mid x_{<t})$, generating each
token conditioned on all preceding tokens in a strict left-to-right order.
This causal factorization aligns naturally with the sequential structure of
language, and next-token prediction provides a uniform self-supervised
objective that scales with data and compute. These properties have driven remarkable progress in large-scale language
modeling~\citep{brown2020gpt3,hoffman2022chinchilla,openai2024gpt4,touvron2023llama,dubey2024llama3,deepseek2024deepseekv3,yang2024qwen3}.
However, the autoregressive factorization constrains generation to proceed
strictly left-to-right, precluding non-autoregressive generation.

\noindent\textbf{Discrete Diffusion Language Models.}
Discrete diffusion language models define a corruption--denoising process
directly over token sequences. D3PMs~\citep{austin2021d3pm} introduced a
general framework covering both absorbing-state and uniform-state discrete
transitions. In the absorbing-state line, tokens are progressively masked
to a special \texttt{[MASK]} token and recovered through iterative
unmasking; MDLM~\citep{sahoo2024mdlm} showed this objective reduces to a
mixture of masked language modeling losses. LLaDA~\citep{nie2025llada}
reported competitive in-context learning performance with LLaMA3 8B, while
Dream~\citep{ye2025dream} improved over previous diffusion language models
on general, mathematical, and coding tasks. BD3-LMs~\citep{arriola2025block} proposed a block-causal
factorization combining autoregressive generation across blocks with masked
diffusion within each block. SEDD~\citep{lou2024sedd} introduced a score
entropy objective that extends score matching to discrete spaces.
Duo~\citep{sahoo2025duo} identified a formal duality between uniform-state
discrete diffusion and Gaussian diffusion, enabling reduced training variance
and few-step generation through discrete consistency distillation. Recent
work has scaled discrete DLMs to reasoning via reinforcement
learning~\citep{zhao2025d1} and to large-scale code
generation~\citep{seedteam2025seeddiffusion}.

In both autoregressive and discrete diffusion models, generation operates
entirely within the discrete token vocabulary. AURORA-LM instead generates
over a continuous latent space, with a learned encoder-decoder mapping
between tokens and high-capacity continuous representations.

\subsection{Continuous Diffusion Language Models}\label{sec:rel-continuous}

\noindent\textbf{Embedding-space Language Models.}
These methods perform denoising in a continuous space derived directly
from token representations, without a separately trained encoder-decoder.
Diffusion-LM~\citep{li2022diffusionlm} first demonstrated controllable
generation by denoising word embedding sequences. Subsequent work, such as
DiffuSeq~\citep{gong2023diffuseq}, extended this formulation to conditional
sequence generation.
PLAID~\citep{gulrajani2023plaid} demonstrated that embedding-space diffusion
can achieve competitive likelihood on standard benchmarks through algorithmic
improvements and scaling analysis. LangFlow~\citep{chen2026langflow} connected
embedding-space diffusion to flow matching via Bregman divergence and
introduced an information-uniform noise schedule, becoming the first continuous
DLM to rival discrete diffusion in perplexity. REPLAID~\citep{yang2026replaid}
established the first scaling law showing that continuous diffusion rivals
discrete DLMs at scale. CoDAR~\citep{shen2026codar} addressed the token
rounding bottleneck with a context-aware autoregressive decoder.

\noindent\textbf{Latent-space Language Models.}
These methods first map text into a continuous latent representation and then model this representation with diffusion or flow matching.
LD4LG~\citep{lovelace2023ld4lg} augments a frozen pretrained
encoder--decoder model with a Perceiver Resampler that compresses encoder
features into a compact, fixed-length latent for diffusion.
TEncDM~\citep{shabalin2024tencdm} instead applies Gaussian diffusion to
full-length contextual representations from a pretrained language encoder,
without compressing the sequence. Building on this encoder-space formulation,
COSMOS~\citep{meshchaninov2025cosmos} introduces a Perceiver-resampler
autoencoder to learn a compressed and smooth latent, while retaining the
TEncDM denoiser architecture and noise schedule.
ELF~\citep{hu2026elf} encodes tokens with a frozen T5 encoder and performs
flow matching in the resulting latent space; the same network serves as both
denoiser and decoder through a shared-weight projection, applying token-level
supervision only at the final step. Cola-DLM~\citep{guo2026cola} learns the
latent via a variational autoencoder with KL regularization, then jointly pretrains the encoder--decoder and the flow-matching model.
TextLDM~\citep{jiang2026textldm} trains a Transformer VAE whose encoder
representations are aligned with hidden states from a frozen Qwen3 model, and
performs flow matching in the resulting latent space.

AURORA-LM follows this factorization but differs in two respects. First, it explicitly constructs a high-capacity, decoder-facing continuous representation with an autoencoder for accurate token recovery, rather than compressing this latent sequence for prior tractability. Second, the autoencoder is frozen before training the denoiser,
fixing the target latent distribution and allowing diffusion-side design
choices to be studied independently of the representation.

\section{The AURORA-LM Framework}\label{sec:method}

\begin{figure}[t]
  \centering
  \includegraphics[width=\linewidth]{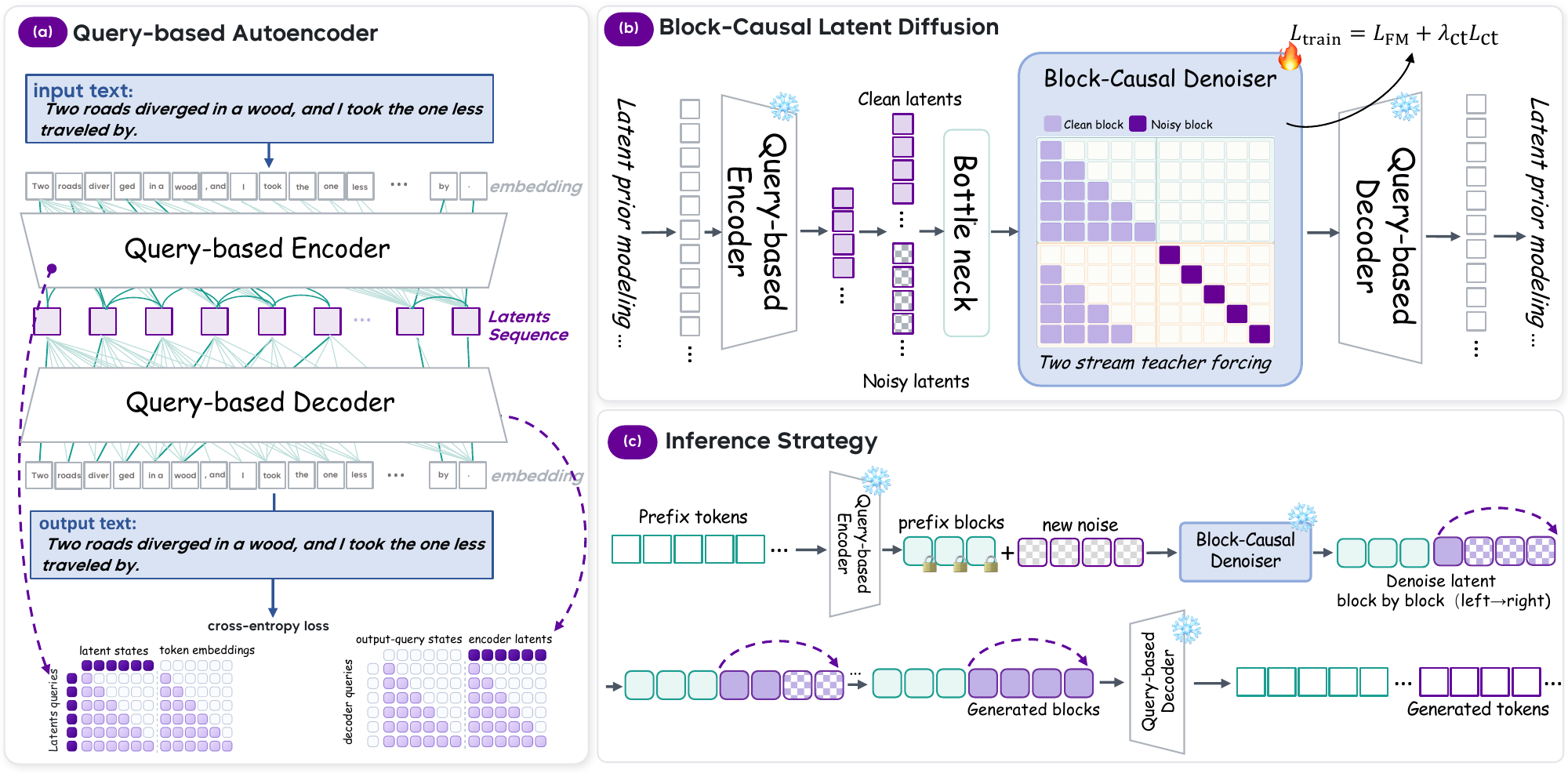}
  \caption{
  \textbf{Overview pipeline of AURORA-LM.}
(a) The Query-based Encoder-Decoder constructs an ordered text latent sequence
from causally expanding input prefixes and maps it back to token logits via a query decoder.
(b) A block-causal denoiser parameterizes the latent prior and is
trained with flow matching. (c) At inference, latent blocks are generated from left to right with KV caching of the completed prefix.}
  \label{fig:method-overview}
\end{figure}

AURORA-LM is organized around the two learning problems identified in Sec.~\ref{sec:intro}: constructing a continuous text representation that can be decoded back into tokens, and modeling its generative distribution. Figure~\ref{fig:method-overview} illustrates the full pipeline.
Specifically, our method consists of a Query-based encoder-decoder and a block-causal denoiser --- a Transformer trained with flow matching to model the continuous text latent distribution.
Given a text sequence, the Query-based Encoder-Decoder organizes it into a high-capacity, causally ordered latent interface between the discrete text and continuous generation. After learning this representation, we freeze the autoencoder, standardize its latent outputs, and train a block-causal denoiser to model their distribution through blockwise flow matching.
Together, these two modules realize continuous language generation during inference by coupling blockwise latent dynamics with a fixed latent-to-token interface.

The remainder of this section follows the same data flow as in Fig.~\ref{fig:method-overview}.
We begin with the flow-matching formulation in Sec.~\ref{sec:prelim}, followed by the construction and decoding of the continuous text representation in Sec.~\ref{sec:ae}. 
Sec.~\ref{sec:prior} introduces the block-causal denoiser for modeling the latent distribution, while Sec.~\ref{sec:prior-training} presents the input bottleneck and training objectives used to learn the full-width latent target.
Sec.~\ref{sec:generation} completes the framework with the blockwise generation and text-decoding procedure. 
We examine the principal design choices governing latent capacity, the modeling of the full-width latent distribution, and efficient blockwise generation in the controlled design analysis of Sec.~\ref{sec:study}.

\subsection{Preliminaries}\label{sec:prelim}

Diffusion models generate data by learning to reverse a gradual transformation from clean samples to noise~\citep{ho2020ddpm,song2021scoresde}. 
In our method, we adopt the flow-matching formulation~\citep{lipman2023flow,liu2023rectified}, which simplifies the procedure by learning a velocity field along straight interpolating paths.
Given a clean sample $x_0$ and Gaussian noise $\varepsilon\sim\mathcal{N}(0,I)$, we define the linear probability path as:

\begin{equation}\label{eq:fm-interpolant}
x_t=(1-t)x_0+t\varepsilon,
\qquad t\in[0,1].
\end{equation}

The path interpolates from data at $t=0$ to noise at $t=1$. A neural network can then learn the reverse dynamics by predicting the path velocity or the clean endpoint $x_0$. 
Although these parameterizations correspond to the same probability path, they lead to different training behavior. 
AURORA-LM adopts clean-endpoint prediction in its final configuration, while Sec.~\ref{sec:study-target} evaluates the alternative parameterizations.

In AURORA-LM, the clean sample is not a token sequence itself. 
It is the continuous sequence produced by the frozen encoder. The general path in Eq.~\eqref{eq:fm-interpolant} is therefore instantiated block by block in Sec.~\ref{sec:prior}, where each clean latent block replaces $x_0$ and is generated conditional on the preceding clean blocks. 
This distinction connects the standard flow-matching formulation to the two parts of our method: the autoencoder first determines what constitutes a clean text latent, and the latent language model subsequently learns its reverse generative process.

\subsection{Continuous Text Latent Construction}
\label{sec:ae}

Our goal is to construct a continuous text representation that serves as an interface between continuous generative modeling and discrete token decoding.
Such a representation must preserve sufficient information for accurate token decoding and organize its latent positions in a causal order suitable for blockwise left-to-right generation.
Rather than inheriting this interface from token embeddings or pretrained encoder features, we construct it using the Query-based Encoder-Decoder illustrated in Fig.~\ref{fig:method-overview}(a).
The encoder constructs successive latent positions from progressively longer token prefixes, while the decoder requires each token prediction to rely on the corresponding latent prefix. 
Together, these encoder and decoder constraints define a prefix-ordered continuous latent sequence trained for token reconstruction. Once the autoencoder is trained and frozen, the block-causal denoiser introduced in Sec.~\ref{sec:prior} models the distribution over these latent sequences.

\noindent \textbf{Latent Sequence.}
Given a token sequence $w_{1:L}$ of length $L$, the encoder produces a continuous latent sequence $z_{\mathrm{enc}}\in\mathbb{R}^{N\times D}$, where $D$ denotes the channel width and $N$ denotes the number of latent positions.
We set $N=\operatorname{round}(cL)$, where $c\in(0,1]$ is the latent retention ratio and $c=1$ denotes no sequence compression, while smaller values produce a shorter latent sequence.
Accordingly, $D$ and $N$ govern complementary aspects of representation
capacity: increasing $D$ gives each latent position more channels for encoding textual information, 
whereas reducing $N$ organizes the same text over fewer latent positions.
We examine this capacity allocation in Sec.~\ref{sec:experiments}. 
The encoder input and decoder output projection share a trainable token embedding matrix $E\in\mathbb{R}^{|V|\times d}$.

\noindent \textbf{Query-based Encoder.}
To construct an ordered latent sequence, we use $N$ latent queries to aggregate a variable-length token sequence into $N$ continuous positions. 
Each query is initialized from the same learnable vector $q\in\mathbb{R}^{D}$, so that $z_{\mathrm{enc},i}^{(0)}=q$ for $i\in\{1,\ldots,N\}$. 
Although initialized identically, the queries are distinguished by the context visible at each position. At layer $\ell$, query $i$ can attend to latent states $z_{\mathrm{enc},1:i}^{(\ell)}$ and to the token prefix 
$E[w_{1:\lceil iL/N\rceil}]$.
We apply RoPE~\citep{su2021roformer} to both sequences before attention to encode their positions.
Specifically, the layer updates are computed by:

\begin{align}
\label{eq:encoder}
    \widetilde z_{\mathrm{enc},i}^{(\ell)} &=z_{\mathrm{enc},i}^{(\ell)}+\mathrm{MHA}^{(\ell)}\!\left(z_{\mathrm{enc},i}^{(\ell)},\;\bigl[z_{\mathrm{enc},1:i}^{(\ell)};E[w_{1:\lceil iL/N\rceil}]\bigr]\right),\notag \\
    z_{\mathrm{enc},i}^{(\ell+1)}
    &=\widetilde z_{\mathrm{enc},i}^{(\ell)}+\mathrm{FFN}^{(\ell)}\!\left(\widetilde z_{\mathrm{enc},i}^{(\ell)}\right).
\end{align}

After the final encoder layer $L_{\mathrm{enc}}$, RMS normalization~\citep{zhang2019rmsnorm} is applied
to produce
$z_{\mathrm{enc},i}=\operatorname{RMSNorm}\!\left(z_{\mathrm{enc},i}^{(L_{\mathrm{enc}})}\right)$.
As the visible token prefix expands with $i$, successive latent positions incorporate progressively more textual context together with the preceding latent states. 
The resulting sequence $z_{\mathrm{enc},1:N}$ therefore follows the left-to-right accumulation of text, aligning each latent prefix with the
textual context used for block-causal conditioning.

\noindent \textbf{Query-based Decoder.}
The encoder above defines how token prefixes are written into the ordered latent sequence. 
We now pair it with a query decoder that defines the reverse mapping from latent prefixes to token logits. 
Because the decoder does not receive the original tokens as input
, it must recover the token content from the learned latent representation.
Specifically, we initialize all $L$ decoder queries from a second shared learnable query $q_{\mathrm{dec}}$, so that $h_j^{(0)}=q_{\mathrm{dec}}$. 
Following the order established by the encoder, output position $j$ can attend to the latent prefix 
$z_{\mathrm{enc},1:\,\lfloor (j-1)N/L\rfloor+1}$
and to the causally available output-query states $h_{1:j}^{(\ell)}$, where $h_j^{(\ell)}$ denotes the decoder hidden state and $\ell$ is the layer index. 
Each decoder layer applies the corresponding masked attention followed by an FFN, and the final hidden state is projected to token logits:

\begin{align}
\label{eq:decoder}
    \widetilde h_j^{(\ell)}
    &=
    h_j^{(\ell)}
    +
    \mathrm{MHA}^{(\ell)}\!\left(
        h_j^{(\ell)},
        \left[
            h_{1:j}^{(\ell)};
            z_{\mathrm{enc},1:\,\lfloor (j-1)N/L\rfloor+1}
        \right]
    \right), \notag \\
    h_j^{(\ell+1)}
    &=
    \widetilde h_j^{(\ell)}
    +
    \mathrm{FFN}^{(\ell)}\!\left(
        \widetilde h_j^{(\ell)}
    \right).
\end{align}

After $L_{\mathrm{dec}}$ layers, the hidden state at output position $j$ is denoted by $h_j=h_j^{(L_{\mathrm{dec}})}$. 
We convert this state into vocabulary logits using the transpose of the token embedding matrix: $o_j=h_jE^\top\in\mathbb{R}^{|V|}$.
The index ranges in Eq.~\eqref{eq:decoder} specify which states are visible to each output position: position $j$ can attend to decoder states $h_{1:j}^{(\ell)}$ and to only the first
$\lfloor (j-1)N/L\rfloor+1$ encoder latents. Recall that encoder latent $i$ is constructed from the token prefix
$w_{1:\lceil iL/N\rceil}$.
Accordingly, when reconstructing this token prefix, the decoder is restricted
to $z_{\mathrm{enc},1:i}$ and cannot access future latents
$z_{\mathrm{enc},i+1:N}$.
This matched visibility preserves the correspondence between token and latent
prefixes under sequence compression.

\noindent \textbf{Autoencoder Training.}
We train the encoder and decoder jointly by matching the reconstructed logits to the original tokens with token-level cross-entropy:
\begin{equation}
\label{eq:ae-loss}
    \mathcal{L}_{\mathrm{AE}} = -\mathbb{E}_{w}\!\left[\frac{1}{L}\sum_{j=1}^{L} \log\mathrm{softmax}(o_j)_{w_j}\right].
\end{equation}
We regularize the autoencoder with token-embedding dropout and latent dropout~\citep{meshchaninov2025cosmos}.
Token-embedding dropout independently zeros the entire input embedding at each token position with probability $p_x$, encouraging the encoder to use surrounding context rather than rely on individual tokens.
Latent dropout independently removes each latent coordinate with probability $p_z$, discouraging the decoder from relying on a small subset of latent features.
We disable both forms of dropout after training, freeze the autoencoder, and extract deterministic latent sequences for the second stage. 
This fixes the decoder-facing representation, allowing the diffusion model to learn its distribution without altering the latent space required for token decoding.

\subsection{Block-Causal Modeling of Continuous Text Latents}
\label{sec:prior}

The autoencoder maps each text sequence to a continuous latent sequence with a fixed causal structure. 
The remaining task is to learn the distribution of these representations so that the model can generate new latent sequences without access to the underlying text.
We use the block-causal denoiser for this purpose: it models the latent distribution and generates new representations that the frozen decoder can map back to tokens.

Before learning the latent distribution, we first apply a fixed per-channel affine normalization to the output $z_{\mathrm{enc}}$ of the frozen encoder.
Let $\mu,s\in\mathbb{R}^{D}$ denote the per-channel mean and standard deviation estimated from valid encoder
outputs. 
We standardize each latent sequence as:
\begin{equation}
\label{eq:standardization}
    z=(z_{\mathrm{enc}}-\mu)\oslash s,
    \qquad
    z_{\mathrm{enc}}=\mu+s\odot z,
\end{equation}
where $\oslash$ and $\odot$ denote element-wise division and multiplication, respectively.
The frozen encoder and fixed normalization induce an empirical distribution $q_{\mathcal E}(z)$ over the standardized latent sequences in the training corpus. 
We train a generative model $p_\theta(z)$ to learn this distribution. 
Combined with the frozen decoder $p_\psi$, the model assigns probability to a token sequence by marginalizing over the continuous latent:
\begin{equation}
\label{eq:latent-text-model}
    p(w) = \int p_\psi\!\left(w\mid\mu+s\odot z\right) p_\theta(z)\,dz.
\end{equation}
Here, $p_\theta$ models the distribution of continuous text latents, while $p_\psi$ decodes sampled latents into token sequences.

\noindent \textbf{Block-wise Factorization.}
Modeling $p_\theta(z)$ as a single joint denoising problem would generate every latent position at
once, leaving the causal organization learned by the autoencoder unused. 
At the other extreme,
generating one latent position at a time would preserve the order but discard most of the parallelism
offered by continuous denoising. 
We therefore adopt an intermediate block-causal formulation that performs left-to-right causal generation across latent blocks while jointly denoising all positions within each block.
We call this intermediate organization a \emph{block-causal factorization}.

Formally, for block size $Q$, the $N$ positions of the standardized latent sequence $z$ are partitioned into $B=\lceil N/Q\rceil$ contiguous blocks $\alpha=(\alpha^{(1)},\ldots,\alpha^{(B)})$, and the prior is factorized as
\begin{equation}
\label{eq:block-factorization}
    p_\theta(\alpha) = \prod_{b=1}^{B} p_\theta\!\left(\alpha^{(b)}\mid \alpha^{(<b)}\right),
    \qquad
    \alpha^{(<b)}=(\alpha^{(1)},\ldots,\alpha^{(b-1)}).
\end{equation}
Because the encoder has organized latent prefixes to represent text prefixes, $\alpha^{(<b)}$ represents the textual context preceding block $b$.
Each factor in Eq.~\eqref{eq:block-factorization} therefore describes a continuation conditioned on an already determined prefix.
The block size $Q$ controls how computation is divided between sequential conditioning across blocks and parallel denoising within a block.

The block-causal factorization determines the context available for generating each block, but it does not yet
specify how the continuous conditional distribution
$p_\theta(\alpha^{(b)}\mid \alpha^{(<b)})$ is learned. 
Since $\alpha^{(b)}$ is a continuous vector rather than a categorical token, we model this distribution with the flow-matching formulation introduced in Sec.~\ref{sec:prelim}, which transforms Gaussian noise into a clean latent block.
For a clean block $\alpha^{(b)}$ and Gaussian noise
$\varepsilon^{(b)}\sim\mathcal{N}(0,I)$, we define the linear path as:
\begin{equation}
\label{eq:fm-linear-path}
    \alpha_t^{(b)} =(1-t)\alpha^{(b)}+t\varepsilon^{(b)}, \qquad \varepsilon^{(b)} \sim \mathcal{N}(0,I),
    \qquad t\in[0,1].
\end{equation}
The path starts from the clean latent block at $t=0$ and reaches Gaussian noise at $t=1$.
Although flow matching is commonly formulated through velocity prediction, the same path can also be learned by predicting its clean endpoint. 
AURORA-LM adopts this parameterization and estimates $\alpha^{(b)}$ from the noisy state $\alpha_t^{(b)}$, the noise level $t$, and the clean prefix $\alpha^{(<b)}$. Alternative prediction targets and loss spaces are compared in Sec.~\ref{sec:study-target}.

This formulation defines the conditional generation objective for each block, but it leaves a key optimization challenge. 
The latent representation retains the channel capacity required for token decoding, so the diffusion model must recover a wide, high-dimensional target from noisy inputs. 
The next subsection introduces the input architecture and training objectives used to address this challenge without reducing the clean latent representation.

\subsection{Learning the Full-Width Latent Distribution}
\label{sec:prior-training}

For each latent block, the diffusion model predicts the full $D$-dimensional clean representation used by the frozen decoder.
Several steps are needed to train this conditional model.
First, the decoder requires a full-width clean block, but the denoiser can process its noisy input through a narrower pathway.
We therefore apply a low-rank bottleneck only to the noisy input while preserving the full-width prediction output.
Second, generation proceeds from block to block, whereas training should evaluate all blocks efficiently.
To this end, we use clean prefixes and a two-stream attention mask to train all block conditionals in parallel.
This produces a flow-matching loss whose noise distribution is calibrated to the latent width.
During generation, however, the diffusion model is applied repeatedly along one sample trajectory. 
We therefore add self-conditioning to pass information between sampling steps, and self-trajectory consistency to align the clean-latent predictions at neighboring states along that trajectory.

\noindent \textbf{Noisy-latent Input Bottleneck.}
The frozen decoder expects a clean latent of width $D$, but the denoiser does not need to process every noisy channel independently at the same width. 
At large noise levels, much of the variation in $\alpha_t^{(b)}$ reflects Gaussian corruption rather than structure useful for recovering the clean block.
We therefore preserve the full-width prediction target while constraining the pathway through which the noisy latent enters the Transformer.
Specifically, each position in $\alpha_t^{(b)}$ is projected to the Transformer hidden width $H$ through two linear layers with an intermediate bottleneck dimension $D_b$:

\begin{equation}
\label{eq:prior-bottleneck}
    W_{\mathrm{in}}=W_{\mathrm{up}}W_{\mathrm{down}},
    \qquad
    W_{\mathrm{down}}\in\mathbb{R}^{D_b\times D},
    \quad
    W_{\mathrm{up}}\in\mathbb{R}^{H\times D_b}.
\end{equation}

When $D_b<\min(D,H)$, the input projection has rank at most $D_b$, requiring the diffusion model to extract a compact representation of the noisy block before Transformer processing.
We apply this bottleneck only to the noisy-input pathway; the output head still predicts the full $D$-dimensional clean block required by the frozen decoder. 
JiT~\citep{li2025jit} and ELF~\citep{hu2026elf} adopt the same separation between a low-rank noisy-input pathway and a full-width clean target.
We study the choice of $D_b$ in Sec.~\ref{sec:study-bottleneck}.

\noindent \textbf{Parallel Blockwise Training.}
At inference time, the diffusion model generates latent blocks sequentially, so block $b$ depends on the previously generated prefix $\alpha^{(<b)}$. 
Reproducing this rollout during training would require multiple model evaluations per sequence.
Instead, we use the clean blocks produced by the
frozen encoder as prefix context and train all block conditionals in parallel.

Parallel training must preserve the same causal dependencies used during generation.
For each training sequence, we retain the clean latent blocks and independently corrupt every target block to obtain its noisy state $\alpha_t^{(b)}$.
We then apply a two-stream attention mask that separates clean prefix context from noisy target states. 
When predicting block $b$, the model can attend to its own noisy positions and the clean prefix $\alpha^{(<b)}$, but not to future clean blocks or the noisy states of other blocks:
\begin{equation}
\label{eq:two-stream-training}
    \hat\alpha_{\theta}^{(b)} = f_\theta\!\left(\alpha_{t}^{(b)},t;\,\alpha^{(<b)}\right).
\end{equation}
Here $f_\theta$ is the block-causal denoiser that parameterizes the latent prior $p_\theta$. 
It predicts the clean block $\hat\alpha_{\theta}^{(b)}$ from the noisy current block $\alpha_t^{(b)}$, the noise level $t$, and the clean prefix $\alpha^{(<b)}$.
This masking scheme evaluates all block conditionals in a single forward pass without violating the block-causal factorization.

We concatenate the predicted blocks to obtain the full latent estimate $\hat{z}_{\theta}
=
\left(
\hat{\alpha}_{\theta}^{(1)},
\ldots,
\hat{\alpha}_{\theta}^{(B)}
\right)$, which approximates the clean target sequence $z$.
Let $\mathcal{J}$ denote the set of valid latent positions included in the objective, and let $\pi(t)$ denote the distribution from which the noise level is sampled. 
We define the clean-endpoint flow-matching loss as:
\begin{equation}
\label{eq:fm-loss}
    \mathcal L_{\mathrm{FM}} = \mathbb E_{\substack{z\sim q_{\mathcal E},\,t\sim\pi,\\
    \varepsilon\sim\mathcal N(0,I)}} \left[\frac{1}{D|\mathcal J|}\sum_{i\in\mathcal J}\left\lVert\hat z_{\theta,i}-z_i\right\rVert_2^2\right].
\end{equation}

The noise-level distribution $\pi(t)$ determines how training samples are allocated along the diffusion path. 
Because the effective signal strength at a given $t$ changes with the latent width $D$, we calibrate $\pi(t)$ to the selected representation width; Sec.~\ref{sec:study-schedule} analyzes this choice.

\noindent \textbf{Self-conditioning.}
The flow-matching objective trains the diffusion model on independently sampled noisy states, whereas inference repeatedly evaluates it along a single denoising trajectory. 
We bridge this gap with self-conditioning~\citep{chen2023analog}, which feeds the clean-latent prediction from the previous sampling step back into the model as an additional input.
Specifically, we expose the model to the same conditioning signal during training. 
With probability $p_{\mathrm{sc}}$, an auxiliary forward pass produces a draft clean-latent estimate, which we detach from the computation graph and provide to the main prediction pass. 
Otherwise, the model receives no draft estimate. 
This training scheme allows the diffusion model to operate both with and without self-conditioning and also supports the guidance procedure described in Sec.~\ref{sec:generation}.

\noindent \textbf{Self-trajectory Consistency.}
Self-conditioning provides the previous clean-latent estimate as an input, but it does not explicitly enforce consistency between predictions at successive sampling states. 
This consistency is important because the estimate at time $t$ determines the updated state at time $t'$, which the model uses for the next prediction. 
Inconsistent estimates of the same clean latent can therefore accumulate errors along the denoising trajectory.
We address this issue with \emph{self-trajectory consistency}, which aligns the clean-latent predictions before and after one update along the model's own sampling trajectory.


\begin{wrapfigure}[12]{r}{0.60\linewidth}
    \vspace{-8pt}
    \centering
    \setlength{\abovecaptionskip}{3pt}
    \setlength{\belowcaptionskip}{0pt}
    \includegraphics[width=\linewidth]{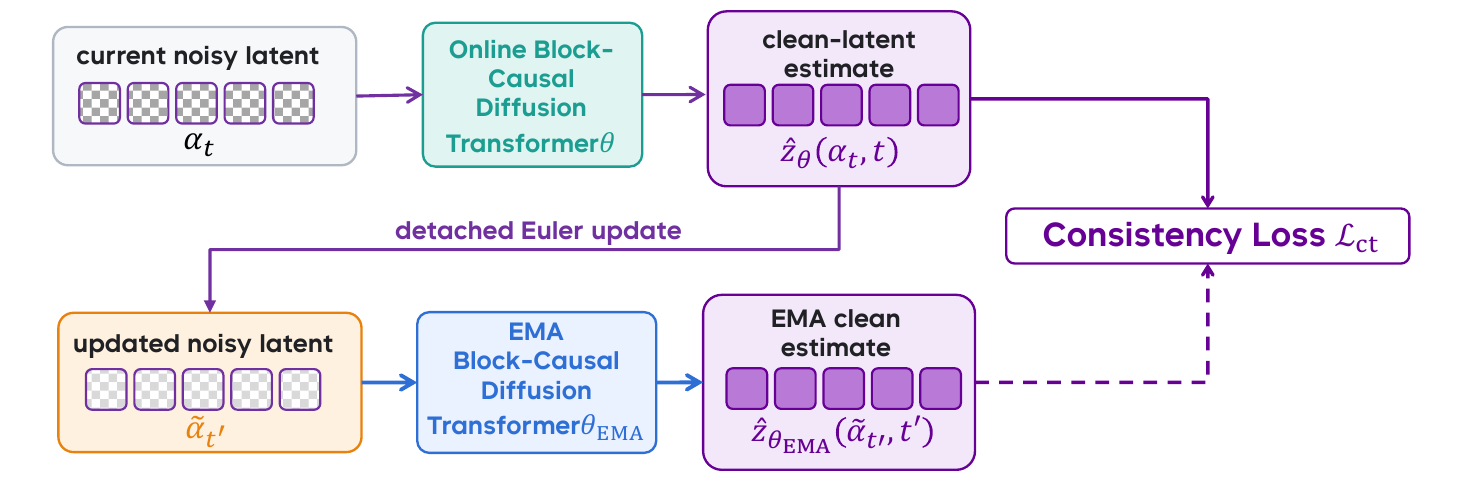}
    \caption{
    \textbf{Overview of self-trajectory consistency.}
    Self-trajectory consistency aligns neighboring clean-latent predictions along a model-induced update.}
    \label{fig:self-trajectory-consistency}
    \vspace{-6pt}
\end{wrapfigure}
We construct the neighboring state using the same schedule parameterization
as sampling. Before training, we fix the training neighbor budget at $S=16$,
which sets a step size of $1/S$ in the schedule coordinate.

For the current noise level $t$ sampled by the flow-matching objective, we
construct the neighboring lower-noise level as
\begin{equation}
\label{eq:consistency-step}
t'=\operatorname{Prev}_{S}(t)<t.
\end{equation}
Here, $\operatorname{Prev}_{S}(t)$ denotes the lower-noise level reached by
taking a step of size $1/S$ from $t$ in the tan-$d$ schedule coordinate
(Appendix~\ref{app:trajectory-consistency}).
Thus, $t'$ is determined by $t$ and the fixed training budget $S$ rather than
sampled independently.

We denote the complete noisy latent sequence by $\alpha_t$ and use the current clean-latent prediction to move it to the lower-noise time $t'$:

\begin{equation}
\label{eq:consistency-neighbor}
\widetilde \alpha_{t'} = \frac{t'}{t}\alpha_t + \left(1-\frac{t'}{t}\right) \operatorname{sg}\!\left(\hat z_\theta(\alpha_t,t)\right),
\end{equation}
where $\operatorname{sg}(\cdot)$ stops gradients through the update. Under the linear path in Eq.~\eqref{eq:fm-linear-path}, Eq.~\eqref{eq:consistency-neighbor} is one Euler step from $t$ to $t'$. 
We use an exponential moving average of the diffusion-model parameters, denoted by
$\theta_{\mathrm{EMA}}$, to provide the target at the updated state. This EMA copy changes more slowly than the current parameters $\theta$, which gives the consistency loss a more stable target.
The current model predicts the clean latent from $(\alpha_t,t)$, while the EMA model predicts it from $(\widetilde \alpha_{t'},t')$. 
We stop gradients through the EMA prediction and minimize:

\begin{equation}
\label{eq:consistency-loss}
    \mathcal L_{\mathrm{ct}}
    =
    \mathbb E_{\substack{
    z\sim q_{\mathcal E},\,
    t\sim\pi,\,
    \varepsilon\sim\mathcal N(0,I)
}}
    \left[
        \frac{1}{D|\mathcal J|}
        \sum_{i\in\mathcal J}
        \left\lVert
            \hat z_\theta(\alpha_t,t)_i
            -
            \operatorname{sg}\!\left(
                \hat z_{\theta_{\mathrm{EMA}}}
                (\widetilde\alpha_{t'},t')_i
            \right)
        \right\rVert_2^2
    \right].
\end{equation}
The complete prior objective is
\begin{equation}\label{eq:total-prior-loss}
\mathcal L_{\mathrm{train}}
=
\mathcal L_{\mathrm{FM}}
+
\lambda_{\mathrm{ct}}\mathcal L_{\mathrm{ct}}.
\end{equation}
The flow-matching objective anchors each prediction to the clean training target, while the consistency objective aligns predictions at neighboring states along the sampling trajectory.
Together, they teach the diffusion model both what clean latent to recover and how to maintain a stable estimate across repeated denoising updates.

\subsection{Latent Generation and Text Decoding}
\label{sec:generation}

AURORA-LM supports both unconditional and prompt-conditioned generation. 
Unconditional generation begins without prefix, while prompt-conditioned generation initializes the prefix by encoding the prompt with the frozen encoder and applying the same latent standardization used during training.
The remaining latent sequence is generated block by block. 
At each stage, the block-causal denoiser conditions on the completed prefix and denoises a newly sampled Gaussian block from $t=1$ to $t=0$.
Its low-rank input pathway processes the noisy block, while self-conditioning propagates the current clean-block estimate across solver steps. 
The self-trajectory consistency objective introduced above further encourages these successive estimates to remain stable along the sampling trajectory. 
Once denoising is complete, the block is appended to the prefix and used to condition the generation of the following block.

\noindent \textbf{Blockwise latent generation.}
During training, the encoder provides the clean prefix for every target block, allowing all block conditionals to be evaluated in parallel. 
At inference time, these prefixes must instead be generated autoregressively.
Following Eq.~\eqref{eq:block-factorization}, the diffusion model generates latent blocks from left to right, with each completed block conditioning the generation of the next.

Before generating block $b$, we collect the previously generated blocks into the prefix $\hat{\alpha}^{(<b)}$ and initialize the current block as Gaussian noise, $\varepsilon^{(b)}\sim\mathcal{N}(0,I)$, at $t=1$. 
The sampler progressively denoises all positions in the block jointly from $t=1$ to $t=0$, conditioning every update on $\hat{\alpha}^{(<b)}$. 
The complete blockwise sampling procedure is written as:
\begin{equation}
\label{eq:block-generation}
    \hat\alpha^{(b)} = \Phi_{\theta,0\leftarrow1}^{(b)} \!\left(\varepsilon_b;\hat\alpha^{(<b)}\right),
    \qquad
    \varepsilon_b\sim\mathcal N(0,I),
\end{equation}
where $\Phi_{\theta,0\leftarrow1}^{(b)}$ denotes the sequence of ODE or SDE solver updates that transforms the initial noise into the clean latent block. 
Starting from an initial latent prefix, each completed block is appended to the context for subsequent generation. 
Repeating this procedure from left to right yields the complete standardized latent sequence:
\[
\hat z=(\hat\alpha^{(1)},\ldots,\hat\alpha^{(B)}).
\]

Unconditional generation begins without prefix and generates the entire latent sequence block by block.
For prompt-conditioned generation, we encode the prompt with the frozen encoder and standardize the resulting latent blocks according to Eq.~\eqref{eq:standardization}. 
These prompt latents initialize the prefix, after which the diffusion model generates only the continuation blocks.

\noindent \textbf{Text decoding.}
The generated sequence $\hat{z}$ lies in the standardized latent space modeled by the denoiser, whereas the frozen decoder operates on the original encoder space.
We therefore recover the decoder-compatible representation by inverting the standardization in Eq.~\eqref{eq:standardization}:
\begin{equation}
\label{eq:generation-inverse-standardization}
    z_{\mathrm{dec}}=\mu+s\odot\hat z.
\end{equation}
The frozen query decoder maps $z_{\mathrm{dec}}$ to the vocabulary logits
$o_{1:L}$ using the decoder mapping described in Sec.~\ref{sec:ae}, from
which the output token sequence is recovered.

\noindent \textbf{Guidance.}
At each solver step, the diffusion model predicts a clean estimate of the current latent block. 
Guidance modifies this estimate by combining two model predictions before updating the noisy state.
For unconditional generation, the clean estimate from the preceding solver step serves as the conditioning signal for SC-CFG. When this signal is available, the model produces one prediction with self-conditioning and one with the self-conditioning input omitted, denoted by $\hat{\alpha}_{\mathrm{sc}}^{(b)}$ and $\hat{\alpha}_{\mathrm{no\text{-}sc}}^{(b)}$, respectively. SC-CFG combines them as:

\begin{equation}
\hat\alpha_{\mathrm{SC\text{-}CFG}}^{(b)}
    =
\hat\alpha_{\mathrm{no\text{-}sc}}^{(b)}
    +
    w_{\mathrm{sc}}
\left(\hat\alpha_{\mathrm{sc}}^{(b)}
    -\hat\alpha_{\mathrm{no\text{-}sc}}^{(b)}
    \right).
\end{equation}
Here, $w_{\mathrm{sc}}=1$ recovers the self-conditioned estimate, while larger values strengthen the direction favored by self-conditioning, providing inference-time control over the trade-off between generation quality and diversity.

For prompt-conditioned generation, standard classifier-free guidance~\citep{ho2022cfg} uses one prediction with the prompt prefix and one without it:
\begin{equation}
\label{eq:cfg}
    \hat\alpha_{\mathrm{CFG}}^{(b)} = \hat\alpha_{\mathrm{uncond}}^{(b)} + w\left( \hat\alpha_{\mathrm{cond}}^{(b)} - \hat\alpha_{\mathrm{uncond}}^{(b)} \right).
\end{equation}
In this case, $w=1$ recovers the prompt-conditioned estimate, while larger values strengthen the effect of the prompt. 
Both guidance rules form linear combinations of clean-latent predictions at inference time and do not alter the training objective. We study the SC-CFG scale in Appendix~\ref{app:research-inference-sweep}.

\section{Experiment}
\label{sec:experiments}

We now evaluate AURORA-LM at three levels.
We begin with controlled short-context ablations that analyze the contribution and rationality of each design choice.
We then compare AURORA-LM-S (with 130M parameters) with existing baselines under matched experimental settings.
To assess the scalability, we further scale up our method and evaluate AURORA-LM-L (with 1.01B parameters) on standard language benchmarks. 
Note that the reported parameter counts include only the block-causal denoiser and exclude the autoencoder.
All experiments are conducted on Ascend NPUs. See detailed configurations in Appendix~\ref{app:protocol-details}.

\noindent \textbf{Metrics.}
We evaluate our AURORA-LM from complementary perspectives:
(1) \textbf{generation perplexity (Gen-PPL)}, computed by GPT-2 Large~\citep{radford2019language}, which measures the fluency of generated text under a fixed external language model;
(2) \textbf{token-unigram entropy}, which measures lexical diversity;
(3) \textbf{MAUVE}~\citep{pillutla2021mauve}, which measures the distributional similarity between generated text and held-out human-written text;
(4) \textbf{ROUGE-1}, \textbf{ROUGE-2}, and \textbf{ROUGE-L}, which evaluate unigram, bigram, and longest-common-subsequence overlap with reference summaries, respectively;
(5) For the scaling-up evaluation, we report task-specific scores and their macro average across the nine benchmarks; task definitions, prompting, answer extraction, and scoring procedures are detailed in Appendix~\ref{app:benchmark-protocol}.

\subsection{Further analysis}
\label{sec:study}

We first conduct controlled ablation studies on OpenWebText~\citep{gokaslan2019openwebtext} with a maximum sequence length of 128 tokens (OWT128). 
Each experiment varies a single design choice while keeping the remaining configuration fixed. 
Unless otherwise specified, we generate 1,000 samples per configuration using a 32-step ODE sampler without guidance and evaluate generation quality with MAUVE against held-out OpenWebText references.

\subsubsection{Latent Capacity}
\label{sec:study-latent}

A key question in continuous language modeling is how much information the latent representation should retain while remaining suitable for generative modeling. 
We examine this question along two dimensions: the channel width $D$, which determines the capacity of each latent position, and sequence compression, which determines how many latent positions are used to represent the text.

\noindent\textbf{Capacity for token recovery.}
We train the autoencoder of Sec.~\ref{sec:ae} with latent widths $D\in\{128,256,512,1024\}$.
To probe the information retained by each representation, we perturb the encoder outputs along $x_{\sigma}=(1-\sigma)z+\sigma\epsilon$, where $\epsilon\sim\mathcal{N}(0,I)$, and decode the corrupted latents directly without denoising.
As shown in Fig.~\ref{fig:study-latent-channel}(a), all configurations reconstruct the original tokens almost perfectly when $\sigma=0$.
Their behavior diverges rapidly as corruption increases: narrow latents lose token-recovery accuracy much earlier, whereas wider latents preserve the information required for exact decoding over a substantially broader corruption range. 
Clean reconstruction alone therefore provides an incomplete measure of representation capacity; robustness to perturbation reveals clear differences between latent widths.

\noindent\textbf{Capacity for generative modeling.}
We next examine whether the additional capacity of wider latents also benefits generation. 
Because latent width changes the appropriate allocation of training noise, we sweep the tan-$d$ schedule shift~\citep{hoogeboom2023simple} over $d\in{1,3,5,7}$ for each $D$ and report the highest MAUVE.
Figure~\ref{fig:study-latent-channel}(b) shows that the best calibrated performance improves consistently with latent width and peaks at $D=1024$. 
Together, the two studies show that wider representations not only retain token-level information more robustly, but also provide a stronger generative target when paired with an appropriately calibrated noise schedule.

\noindent\textbf{Sequence compression.}
Sequence compression offers a complementary means of reducing generation cost.
Shorter latent sequences require the denoiser to generate fewer positions and, for a fixed block size, complete fewer blockwise denoising stages. 
Because the causal prefix structure is preserved under compression, each retained position simply covers a wider token span. 
We fix $D=1024$ and retain $100\%$, $90\%$, $80\%$, or $70\%$ of the latent positions for analysis. 
Figure~\ref{fig:study-latent-channel}(c) shows a clear quality-efficiency trade-off: retaining $90\%$ of the positions produces only a small reduction in MAUVE, while stronger compression leads to a steady decline.
We therefore use the full latent sequence ($c=1$) with $D=1024$ in subsequent ablations to avoid introducing compression-related degradation into the reference configuration.
The setting $c=0.9$ nevertheless provides a practical alternative when generation efficiency is prioritized.

\begin{center}
  \begin{minipage}{0.98\linewidth}
  \centering
  \includegraphics[width=\linewidth]{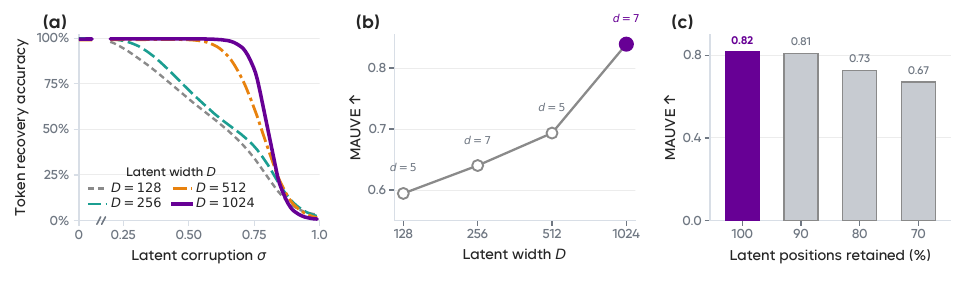}
  \captionof{figure}{
  \textbf{Evaluations of latent capacity.}
  \uline{Latent capacity improves decoder robustness and MAUVE.}
  (a) Wider channels preserve token-recovery accuracy under stronger latent
  corruption; the broken horizontal axis compresses
  $\sigma\in(0.05,0.20)$, where all widths remain near perfect recovery.
  (b) For each $D$, we report the highest MAUVE over the same
  $d\in\{1,3,5,7\}$ search; annotations indicate the selected shift.
  (c) MAUVE over the practical sequence-compression range, expressed as the
  percentage of latent positions retained; the full stress-test sweep is
  reported in Appendix~\ref{app:full-diagnostics}.}
  \label{fig:study-latent-channel}
  \end{minipage}
\end{center}

\subsubsection{Modeling the Full-Width Latent Distribution}
\label{sec:study-prior}

The preceding analysis identifies $D=1024$ as the preferred latent width. 
We now examine three design choices introduced in Sec.~\ref{sec:prior-training} for modeling this full-width target with the denoiser:
the width of the noisy-input pathway, the allocation of flow-matching supervision across noise levels, and the prediction target and loss space.

\noindent\textbf{Noisy-input bottleneck.}
\phantomsection\label{sec:study-bottleneck} The clean prediction target retains full width for accurate token recovery, but the noisy-input pathway serves a different role: it determines how much of the corrupted state the Transformer reads at each denoising step, and its width can be treated as an independent prior-side variable. 
We therefore fix the clean target at $D=1024$ and vary only the bottleneck width $D_b$ of the noisy-input pathway in Eq.~\eqref{eq:prior-bottleneck}. 
As shown in Fig.~\ref{fig:study-bottleneck-schedule}(a), $D_b=128$ achieves the highest mean MAUVE across ODE samplers with 16--64 steps.
A narrower bottleneck with $D_b=32$ restricts the input excessively, whereas increasing the width beyond 128 provides no consistent improvement. 
These results show that the model can retain the full capacity required for clean-latent prediction without processing the corrupted input at full width.
We use $D_b=128$ in the remaining experiments.

\noindent\textbf{Noise allocation.}
\phantomsection\label{sec:study-schedule} 
The noise-level distribution determines how flow-matching supervision is allocated along the corruption path. 
Because the preceding width sweep shows that the preferred noise allocation varies with latent width, the schedule must be calibrated to the selected target representation.
We study this relationship using the tan-$d$ parameterization~\citep{hoogeboom2023simple},

\begin{equation}
    \sigma(\tau;d) = \frac{d\tan(\pi\tau/2)}{1+d\tan(\pi\tau/2)},
    \label{eq:study-tand-map}
\end{equation}
where larger values of $d$ assign more training mass to highly corrupted states. 
We also evaluate logit-normal sampling~\citep{esser2024sd3}, which draws $\ell\sim\mathcal{N}(\mu,s^2)$ and sets $\sigma=\operatorname{sigmoid}(\ell)$.

To compare the parameterizations on a common axis, we vary $d$ for tan-$d$ and $\mu$ for logit-normal with $s\in\{0.5,1.0\}$, and express each setting by its high-noise mass $m_{0.7}=\Pr(\sigma>0.7)$.
Fig.~\ref{fig:study-bottleneck-schedule}(b) shows that MAUVE consistently improves as the schedules allocate more supervision to high-noise states. 
This trend holds across all three sweep families, indicating that modeling wider latent targets benefits from stronger high-noise calibration rather than from a particular schedule parameterization.
We use $d=5$ as the shared reference setting for the controlled component ablations and adopt tan-$d$ with $d=7$ in the final system.

\begin{center}
  \begin{minipage}{0.98\linewidth}
  \centering
  \includegraphics[width=\linewidth]{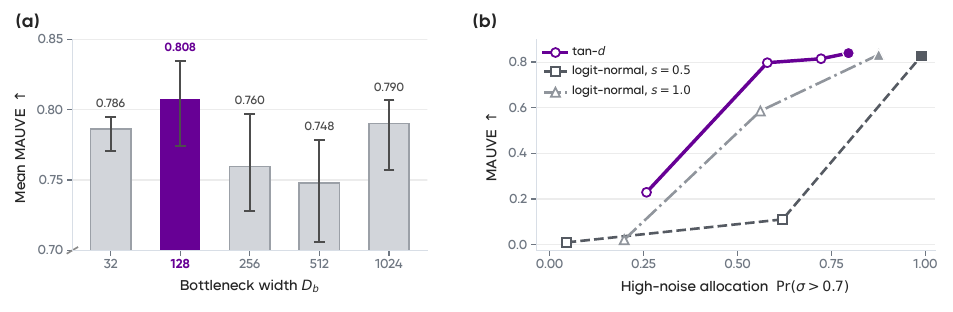}
  \captionof{figure}{\textbf{Evaluations of noisy-input bottleneck and noise allocation.}
  \uline{Moderate noisy-input width and high-noise calibration improve
  modeling of the full-width latent.}
  (a) Bars show mean MAUVE over 16-, 32-, and 64-step ODE sampling;
whiskers show the corresponding min--max range. The clean prediction target remains $D=1024$ while only $D_b$ varies.
  (b) Tan-$d$ sweeps $d\in\{1,3,5,7\}$ and logit-normal sweeps
  $\mu\in\{0,1,2\}$ at $s\in\{0.5,1.0\}$; all settings are expressed by the
  common high-noise mass $m_{0.7}=\Pr(\sigma>0.7)$; the corresponding noise
  distributions are shown in Appendix~\ref{app:schedule-distributions}.
  MAUVE increases with high-noise allocation across all three sweeps; filled
  markers denote the largest tested value ($d=7$ and $\mu=2$).}
  \label{fig:study-bottleneck-schedule}
  \end{minipage}
\end{center}

\begin{wraptable}{r}{0.43\linewidth}
  \centering
  \small
  \vspace{-0.9em}
  \caption{\textbf{Analysis of target and loss space.} Direct clean-latent
regression yields the highest MAUVE under both noisy-input widths.}
  \label{tab:study-target-loss}
  \vspace{0.35em}
  \setlength{\tabcolsep}{3.5pt}
  \renewcommand{\arraystretch}{1.08}
  \begin{tabular}{ccrr}
    \toprule
    & & \multicolumn{2}{c}{\textbf{Noisy-input width}} \\
    \cmidrule(lr){3-4}
    \textbf{Target} & \textbf{Loss} & $D_b=128$ & $D_b=1024$ \\
    \midrule
    \multirow{2}{*}{$x_0$}
      & $x_0$ & \cellcolor[RGB]{232,232,255}\textbf{0.815}
              & \cellcolor[RGB]{232,232,255}\textbf{0.807} \\
      & $v$   & 0.059 & 0.017 \\
    \multirow{2}{*}{$v$}
      & $x_0$ & 0.729 & 0.325 \\
      & $v$   & 0.653 & 0.301 \\
    \bottomrule
  \end{tabular}
  \vspace{-0.5em}
\end{wraptable}

\noindent\textbf{Prediction target and loss space.}
\phantomsection\label{sec:study-target}
After selecting the noise allocation, we separately examine what the network
predicts and the space in which we measure its regression error. We compare the
four combinations formed by clean-latent ($x_0$) or velocity ($v$)
prediction and losses computed in either $x_0$ or $v$ space. We repeat this
comparison with the selected low-rank noisy-input projection ($D_b=128$) and
with $D_b=1024$, which removes the intermediate low-rank bottleneck.
Table~\ref{tab:study-target-loss} shows that predicting $x_0$ with an
$x_0$-space loss achieves the highest MAUVE in both settings.

Computing a $v$-space loss on an $x_0$ prediction performs poorly because the
coordinate conversion weights the clean-latent error by $1/\sigma^2$. This
weighting concentrates supervision on low-noise states and conflicts with the
high-noise allocation identified above (Appendix~\ref{app:target-loss}).
$v$-prediction likewise yields lower MAUVE in our setting: although $v$ can be
converted to a clean-latent estimate, it requires the model to regress a
noise-dependent displacement rather than directly the decoder-facing clean
latent, consistent with prior evidence that direct clean-data prediction can be
advantageous in high-dimensional representation spaces~\citep{li2025jit,hu2026elf}.
This MAUVE degradation becomes more pronounced when the noisy-input pathway is
widened from $D_b=128$ to $D_b=1024$, whereas $x_0$ prediction with an
$x_0$-space loss changes little. We therefore adopt $x_0$ prediction with an $x_0$-space loss, directly
supervising the representation that the frozen decoder maps back to tokens.

\subsubsection{Efficient Blockwise Generation}
\label{sec:study-efficiency}

\noindent\textbf{Few-step sampling.}
Flow matching supervises individual noisy states independently, but generation chains these states along the model's own trajectory. 
Prediction errors at neighboring states can therefore propagate across solver steps, especially when only a few updates are available.
Our proposed self-trajectory consistency mitigates this mismatch by aligning clean-latent predictions at successive states along the model-induced trajectory.
As shown in
Fig.~\ref{fig:study-trajectory-block}(a), it improves MAUVE at every tested
step budget, with the largest gains in the few-step regime.
The advantage gradually narrows as the solver uses more steps and the trajectory becomes more finely discretized.
These results confirm that trajectory
consistency is necessary to realize the few-step generation quality of the
block-causal denoiser.

\noindent\textbf{Block granularity.}
The block size $Q$ controls the balance between left-to-right conditioning across blocks and parallel denoising within each block.
We train separate models with $Q\in\{4,8,16,32,64\}$ and evaluate them with 32-step ODE sampling and no guidance. 
As shown in Fig.~\ref{fig:study-trajectory-block}(b), MAUVE decreases from $0.909$ at $Q=4$ to $0.631$ at $Q=64$.
Smaller blocks provide each target block with a more complete generated prefix, improving conditional generation at the cost of more sequential stages.
Larger blocks increase within-block parallelism but weaken the amount of preceding context available to each generation step.
We choose $Q=16$ as a balanced operating point, achieving a MAUVE of $0.815$ while reducing the number of sequential block-generation stages from 32 to 8 relative to $Q=4$.

\begin{center}
  \begin{minipage}{0.98\linewidth}
  \centering
  \includegraphics[width=\linewidth]{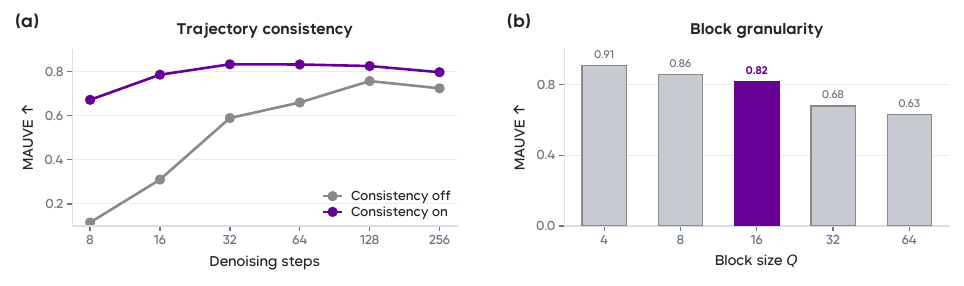}
  \captionof{figure}{\textbf{Evaluation over blockwise generation.} \uline{Self-trajectory consistency improves few-step sampling, while smaller blocks trade parallelism for higher generation quality.}
(a) MAUVE as a function of denoising-step budget, with and without self-trajectory consistency.
Consistency improves MAUVE at every budget, with the largest gains in the few-step regime.
(b) MAUVE for separately trained models across block sizes $Q\in\{4,8,16,32,64\}$, evaluated with 32-step ODE sampling.
Smaller blocks provide more sequential conditioning stages and yield higher MAUVE; the filled marker denotes the selected $Q=16$ operating point.}
  \label{fig:study-trajectory-block}
  \end{minipage}
\end{center}

\subsection{System-Level Comparison}
\label{sec:scaling}

We then evaluate AURORA-LM-S on OpenWebText~\citep{gokaslan2019openwebtext} free generation and XSum~\citep{narayan2018xsum} conditional summarization.
To enable controlled comparison across generation paradigms, we evaluate released autoregressive, discrete-diffusion, and continuous-generation models under a common evaluation protocol where available.

\noindent\textbf{Unconditional generation.}
We train AURORA-LM-S on OpenWebText using sequences of 1,024 tokens.
During evaluation, each model generates 1,000 samples without an input prompt. 
We report generation perplexity (Gen-PPL), computed by GPT-2 Large~\citep{radford2019language} as a fixed external evaluator, token-unigram entropy for lexical diversity, and MAUVE~\citep{pillutla2021mauve} for distributional similarity between generated samples and held-out OpenWebText passages.

\noindent\textbf{Conditional summarization.}
We train and evaluate AURORA-LM-S on XSum~\citep{narayan2018xsum}.
AURORA-LM-S conditions on the source article and generates its summary.
We measure summarization quality using ROUGE-1, ROUGE-2, and ROUGE-L~\citep{lin2004rouge}.

\noindent\textbf{Baselines.}
We reproduce baselines where published implementations or released checkpoints
are available and evaluate them through a common pipeline. The corresponding
tables mark results quoted directly from prior work with $^\dagger$.

\subsubsection{Unconditional Generation}
\label{sec:scale-uncond}

We first evaluate whether the selected configuration is competitive against existing
generation paradigms on long-context free generation.
Specifically, we compare AURORA-LM-S against AR, Duo and Duo-distilled~\citep{sahoo2025duo}, SEDD~\citep{lou2024sedd}, MDLM~\citep{sahoo2024mdlm}, and ELF-B~\citep{hu2026elf}, covering autoregressive, discrete-diffusion, and
continuous-flow generation. 
AURORA-LM-S uses 64-step SDE-DPM++~\citep{lu2022dpmsolverpp} sampling with an SC-CFG scale of 4; sampling configurations for all other systems are provided in Appendix~\ref{app:research-inference-sweep}.

\begin{table}[t]
  \centering
  \scriptsize
  \setlength{\tabcolsep}{2.0pt}
  \caption{\textbf{Unconditional generation on OpenWebText (1,024 tokens).} Scores are averaged over five generation seeds, with 1,000 samples per seed.
  Gen-PPL uses GPT-2 Large as a fixed external scorer; MAUVE is computed against held-out OpenWebText references. All
  rows are in-house reproductions. Parameter counts follow model-specific conventions (Table~\ref{tab:model_arch}): AURORA counts the denoiser only, whereas ELF includes its output head. Detailed sampler and guidance
  configurations are in Appendix~\ref{app:research-inference-sweep}.
  \label{tab:scale-research}}
  \begin{tabular}{lcccc}
    \toprule
    \textbf{Model} & \textbf{Params.} & \textbf{Gen-PPL $\downarrow$} &
    \textbf{Entropy $\uparrow$} & \textbf{MAUVE $\uparrow$} \\
    \midrule
    AR & 85M & 39.40 & 5.605 & 0.851 \\
    Duo & 92M & 86.57 & 5.566 & 0.704 \\
    Duo-distilled & 92M & 78.22 & 5.574 & 0.715 \\
    SEDD & 92M & 119.82 & 5.644 & 0.693 \\
    MDLM & 92M & 121.36 & \textbf{5.664} & 0.668 \\
    ELF-B & 105M & 24.11 & 5.155 & ---$^{*}$ \\
    \midrule
    \rowcolor[RGB]{232,232,255}
    \textbf{AURORA-LM-S} & 130M & \textbf{23.56} & 5.241 & \textbf{0.890} \\
    \bottomrule
  \end{tabular}
  \par\smallskip
  \begin{minipage}{\linewidth}
  \scriptsize
  $^{*}$ELF-B's MAUVE uses different reference data and truncation; its result is reported separately in Appendix~\ref{app:elf-mauve-protocol}.
  \end{minipage}
\end{table}

As shown in Table~\ref{tab:scale-research}, AURORA-LM achieves the lowest Gen-PPL (23.56) among the evaluated models and the highest MAUVE (0.890) under the shared evaluation protocol.
These results demonstrate that generating a learned continuous text representation provides strong long-context generation quality and compares favorably with both discrete and existing continuous formulations.

\subsubsection{Conditional Generation}
\label{sec:scale-cond}

We next test whether the selected configuration generalizes to conditional generation. We evaluate AURORA-LM-S on XSum~\citep{narayan2018xsum}, reporting ROUGE-1, ROUGE-2, and ROUGE-L. 
AURORA-LM-S uses 32-step ODE sampler with classifier-free guidance at scale $w=2.0$; generation details are described
in Appendix~\ref{app:generation-protocols}.

\begin{table}[t]
  \centering
  \scriptsize
  \setlength{\tabcolsep}{4.0pt}
  \caption{\textbf{XSum conditional-generation comparison.} AURORA-LM-S is
  evaluated under the matched protocol described in
  Appendix~\ref{app:generation-protocols}. $^\dagger$Values collected by
  \citet{hu2026elf}.\label{tab:scale-xsum}}
  \begin{tabular}{lccc}
    \toprule
    \textbf{Model} & \textbf{ROUGE-1 $\uparrow$} &
    \textbf{ROUGE-2 $\uparrow$} & \textbf{ROUGE-L $\uparrow$} \\
    \midrule
    ELF-B$^\dagger$ & 36.0 & 12.2 & 27.8 \\
    AR$^\dagger$ & 30.5 & 10.2 & 24.4 \\
    MDLM$^\dagger$ & 33.4 & 11.6 & 25.8 \\
    Duo$^\dagger$ & 31.4 & 10.1 & 25.0 \\
    E2D2$^\dagger$ & 28.4 & 8.3 & 22.0 \\
    SeqDiffuSeq$^\dagger$~\citep{yuan2024seqdiffuseq} & 19.3 & 1.7 & 14.1 \\
    \midrule
    \rowcolor[RGB]{232,232,255}
    \textbf{AURORA-LM-S} & \textbf{36.6} & \textbf{13.4} & \textbf{28.9} \\
    \bottomrule
  \end{tabular}
\end{table}

As shown in Table~\ref{tab:scale-xsum}, AURORA-LM achieves the best ROUGE scores across all three metrics, outperforming ELF-B and all other evaluated systems. 
Together with the unconditional generation results above, these findings show that the same continuous-latent formulation performs strongly in both unconditional generation and prompt-conditioned generation tasks.

\subsection{Scaling Evaluation}
\label{sec:scale-benchmark}

The preceding experiments establish AURORA-LM at the 130M scale on generation-focused evaluations. 
We next examine whether its advantages persist when the diffusion model scales to approximately 1B parameters and is evaluated across a broader range of language capabilities. 
AURORA-LM-L is trained on open-source pretraining data with an estimated total compute of approximately 1,500 EFLOPs.  Appendix~\ref{app:protocol-details} provides the full training configuration.

We compare AURORA-LM-L with Cola-DLM~\citep{guo2026cola}, a publicly released latent-diffusion language model whose diffusion Transformer contains approximately 1.8B parameters. The released Cola-DLM checkpoint corresponds
to the approximately 2,000-EFLOP endpoint of its reported scaling curve. We evaluate its released checkpoint using the official inference implementation and default sampling protocol. AURORA-LM-L uses 16-step ODE
sampler with classifier-free guidance at scale $w=3.0$; full sampling details are provided in Appendix~\ref{app:generation-protocols}. The benchmark suite contains nine public tasks covering commonsense reasoning, factual knowledge, text continuation, reading comprehension, and question answering. We use two-shot prompting for tasks that support demonstrations; Appendix~\ref{app:benchmark-protocol} details the evaluation settings and the meaning of each metric.

\begin{table}[h]
  \centering
  \scriptsize
  \setlength{\tabcolsep}{2.5pt}
  \caption{\textbf{Benchmark evaluation.} All entries are percentages; Avg is the macro
  average over the nine tasks. Cola-DLM uses an approximately 1.8B-parameter
  DiT; AURORA-LM-L uses an approximately 1B-parameter block-causal denoiser
  Transformer. Higher values indicate better performance for all metrics. \label{tab:scale-benchmark}}
  \resizebox{\linewidth}{!}{%
  \begin{tabular}{lcccccccccc}
    \toprule
    \textbf{Model} & \textbf{Avg $\uparrow$} &
    \textbf{MMLU} & \textbf{ARC-C} & \textbf{OBQA} &
    \textbf{HellaSwag} & \textbf{WinoGrande} & \textbf{StoryCloze} &
    \textbf{SIQA} & \textbf{RACE} & \textbf{SQuAD EM} \\
    \midrule
    \rowcolor[RGB]{232,232,255}
    \textbf{AURORA-LM-L (1B)} &
    \textbf{32.6} & \textbf{22.2} & \textbf{21.2} & \textbf{27.8} &
    \textbf{18.4} & \textbf{50.3} & \textbf{54.8} &
    \textbf{30.2} & \textbf{30.6} & \textbf{38.2} \\
    Cola-DLM (1.8B) &
    25.1 & 19.6 & 20.6 & 24.2 &
    5.7 & 45.3 & 33.8 &
    26.8 & 24.2 & 25.7 \\
    \bottomrule
  \end{tabular}}
\end{table}

Table~\ref{tab:scale-benchmark} shows that AURORA-LM-L achieves a macro average of 32.6, compared with 25.1 for Cola-DLM, and performs better on all nine benchmarks.
These results indicate that the gains of AURORA-LM extend beyond the controlled research-scale setting: despite using a smaller diffusion Transformer, AURORA-LM-L consistently outperforms the larger publicly released latent-diffusion model across diverse language tasks.

\section{Conclusion}
\label{sec:conclusion}

We presented AURORA-LM, a continuous-latent diffusion language model that decouples representation learning from distribution modeling.
We first use a Query-based Encoder-Decoder to construct a high-capacity latent sequence that supports accurate token recovery and follows a causal order suitable for blockwise left-to-right generation.
We then freeze the autoencoder and train a block-causal diffusion Transformer to model the resulting full-width latent distribution through flow matching.
To make this distribution easier to learn without reducing the information available for decoding, we restrict only the noisy-input pathway with a low-rank bottleneck, calibrate the noise-level distribution to the latent width, and introduce self-trajectory consistency to align predictions across successive denoising states.

Our controlled studies validate the main design choices of AURORA-LM. 
Wider latent representations retain token-level information more robustly under corruption and yield stronger generation quality after noise calibration.
A moderate noisy-input bottleneck improves latent modeling while preserving the full-width clean prediction target. 
The experiments also support clean-latent prediction, high-noise training allocation, self-trajectory consistency, and block-causal generation.
Under matched training and evaluation protocols, AURORA-LM achieves strong results on both OpenWebText free generation and XSum conditional summarization. 
We further scale up the diffusion model to 1B parameters and outperform a larger publicly released latent-diffusion language model across a diverse set of language benchmarks.

These results show that a high-capacity, causally structured, and decodable continuous text representation can serve as an effective interface between diffusion-based generation and discrete token decoding. 
We believe this formulation provides a promising basis for scaling continuous language models to longer contexts and broader language capabilities, as well as for developing unified generative models that operate across language and other continuous modalities.




\clearpage
\bibliographystyle{plainnat}
\bibliography{references}

\begin{thebibliography}{67}
\providecommand{\natexlab}[1]{#1}
\providecommand{\url}[1]{\texttt{#1}}
\expandafter\ifx\csname urlstyle\endcsname\relax
  \providecommand{\doi}[1]{doi: #1}\else
  \providecommand{\doi}{doi: \begingroup \urlstyle{rm}\Url}\fi

\bibitem[Albergo and Vanden-Eijnden(2023)]{albergo2023stochastic}
Michael~S. Albergo and Eric Vanden-Eijnden.
\newblock Building normalizing flows with stochastic interpolants.
\newblock In \emph{International Conference on Learning Representations
  (ICLR)}, 2023.

\bibitem[Arriola et~al.(2025)Arriola, Gokaslan, Chiu, Yang, Qi, Han, Sahoo, and
  Kuleshov]{arriola2025block}
Marianne Arriola, Aaron Gokaslan, Justin~T. Chiu, Zhihan Yang, Zhixuan Qi,
  Jiaqi Han, Subham~Sekhar Sahoo, and Volodymyr Kuleshov.
\newblock Block diffusion: Interpolating between autoregressive and diffusion
  language models.
\newblock In \emph{International Conference on Learning Representations
  (ICLR)}, 2025.

\bibitem[Austin et~al.(2021)Austin, Johnson, Ho, Tarlow, and van~den
  Berg]{austin2021d3pm}
Jacob Austin, Daniel~D. Johnson, Jonathan Ho, Daniel Tarlow, and Rianne van~den
  Berg.
\newblock Structured denoising diffusion models in discrete state-spaces.
\newblock In \emph{Advances in Neural Information Processing Systems
  (NeurIPS)}, 2021.

\bibitem[Brown et~al.(2020)Brown, Mann, Ryder, Subbiah, et~al.]{brown2020gpt3}
Tom~B. Brown, Benjamin Mann, Nick Ryder, Melanie Subbiah, et~al.
\newblock Language models are few-shot learners.
\newblock In \emph{Advances in Neural Information Processing Systems
  (NeurIPS)}, 2020.

\bibitem[{Chameleon Team}(2024)]{chameleon2024chameleon}
{Chameleon Team}.
\newblock Chameleon: Mixed-modal early-fusion foundation models.
\newblock \emph{arXiv preprint arXiv:2405.09818}, 2024.

\bibitem[Chen et~al.(2023)Chen, Zhang, and Hinton]{chen2023analog}
Ting Chen, Ruixiang Zhang, and Geoffrey Hinton.
\newblock Analog bits: Generating discrete data using diffusion models with
  self-conditioning.
\newblock In \emph{International Conference on Learning Representations
  (ICLR)}, 2023.

\bibitem[Chen et~al.(2026)Chen, Liang, Sui, Guo, Cheng, You, and
  Liu]{chen2026langflow}
Yuxin Chen, Chumeng Liang, Hangke Sui, Ruihan Guo, Chaoran Cheng, Jiaxuan You,
  and Ge~Liu.
\newblock Langflow: Continuous diffusion rivals discrete in language modeling.
\newblock \emph{arXiv preprint arXiv:2604.11748}, 2026.

\bibitem[Clark et~al.(2018)Clark, Cowhey, Etzioni, Khot, Sabharwal, Schoenick,
  and Tafjord]{clark2018arc}
Peter Clark, Isaac Cowhey, Oren Etzioni, Tushar Khot, Ashish Sabharwal, Carissa
  Schoenick, and Oyvind Tafjord.
\newblock Think you have solved question answering? {T}ry {ARC}, the {AI2}
  reasoning challenge.
\newblock \emph{arXiv preprint arXiv:1803.05457}, 2018.

\bibitem[DeepSeek-AI et~al.(2024)DeepSeek-AI, Liu, Feng, Wang, Wang, Liu, Zhao,
  Deng, Ruan, Dai, Guo, et~al.]{deepseek2024deepseekv3}
DeepSeek-AI, Aixin Liu, Bei Feng, Bin Wang, Bingxuan Wang, Bo~Liu, Chenggang
  Zhao, Chengqi Deng, Chong Ruan, Damai Dai, Daya Guo, et~al.
\newblock {DeepSeek-V3} technical report.
\newblock \emph{arXiv preprint arXiv:2412.19437}, 2024.

\bibitem[Dubey et~al.(2024)Dubey, Jauhri, Pandey, Kadian,
  et~al.]{dubey2024llama3}
Abhimanyu Dubey, Abhinav Jauhri, Abhinav Pandey, Abhishek Kadian, et~al.
\newblock The llama 3 herd of models.
\newblock \emph{arXiv preprint arXiv:2407.21783}, 2024.

\bibitem[Esser et~al.(2024)Esser, Kulal, Blattmann, Entezari, M{\"u}ller,
  Saini, Levi, Lorenz, Sauer, Boesel, Podell, Dockhorn, English, Lacey,
  Goodwin, Marek, and Rombach]{esser2024sd3}
Patrick Esser, Sumith Kulal, Andreas Blattmann, Rahim Entezari, Jonas
  M{\"u}ller, Harry Saini, Yam Levi, Dominik Lorenz, Axel Sauer, Frederic
  Boesel, Dustin Podell, Tim Dockhorn, Zion English, Kyle Lacey, Alex Goodwin,
  Yannik Marek, and Robin Rombach.
\newblock Scaling rectified flow transformers for high-resolution image
  synthesis.
\newblock In \emph{International Conference on Machine Learning (ICML)}, 2024.

\bibitem[Gokaslan and Cohen(2019)]{gokaslan2019openwebtext}
Aaron Gokaslan and Vanya Cohen.
\newblock Openwebtext corpus.
\newblock \url{https://skylion007.github.io/OpenWebTextCorpus/}, 2019.

\bibitem[Gong et~al.(2023)Gong, Li, Feng, Wu, and Kong]{gong2023diffuseq}
Shansan Gong, Mukai Li, Jiangtao Feng, Zhiyong Wu, and Lingpeng Kong.
\newblock Diffuseq: Sequence to sequence text generation with diffusion models.
\newblock In \emph{International Conference on Learning Representations
  (ICLR)}, 2023.

\bibitem[Gulrajani and Hashimoto(2023)]{gulrajani2023plaid}
Ishaan Gulrajani and Tatsunori~B. Hashimoto.
\newblock Likelihood-based diffusion language models.
\newblock In \emph{Advances in Neural Information Processing Systems},
  volume~36, 2023.

\bibitem[Guo et~al.(2026)Guo, Zhao, Zhao, Nie, Zhu, Guo, Wang, Yang, Zhao, Wei,
  and Zeng]{guo2026cola}
Hongcan Guo, Qinyu Zhao, Yian Zhao, Shen Nie, Rui Zhu, Qiushan Guo, Feng Wang,
  Tao Yang, Hengshuang Zhao, Guoqiang Wei, and Yan Zeng.
\newblock Continuous latent diffusion language model.
\newblock \emph{arXiv preprint arXiv:2605.06548}, 2026.

\bibitem[Hendrycks et~al.(2021)Hendrycks, Burns, Basart, Zou, Mazeika, Song,
  and Steinhardt]{hendrycks2021mmlu}
Dan Hendrycks, Collin Burns, Steven Basart, Andy Zou, Mantas Mazeika, Dawn
  Song, and Jacob Steinhardt.
\newblock Measuring massive multitask language understanding.
\newblock In \emph{International Conference on Learning Representations}, 2021.

\bibitem[Henry et~al.(2020)Henry, Dachapally, Pawar, and Chen]{henry2020qknorm}
Alex Henry, Prudhvi~Raj Dachapally, Shubham~Shantaram Pawar, and Yuxuan Chen.
\newblock Query-key normalization for transformers.
\newblock In \emph{Findings of the Association for Computational Linguistics:
  EMNLP 2020}, pages 4246--4253. Association for Computational Linguistics,
  2020.
\newblock \doi{10.18653/v1/2020.findings-emnlp.379}.

\bibitem[Ho and Salimans(2022)]{ho2022cfg}
Jonathan Ho and Tim Salimans.
\newblock Classifier-free diffusion guidance.
\newblock \emph{arXiv preprint arXiv:2207.12598}, 2022.

\bibitem[Ho et~al.(2020)Ho, Jain, and Abbeel]{ho2020ddpm}
Jonathan Ho, Ajay Jain, and Pieter Abbeel.
\newblock Denoising diffusion probabilistic models.
\newblock In \emph{Advances in Neural Information Processing Systems
  (NeurIPS)}, 2020.

\bibitem[Ho et~al.(2022)Ho, Salimans, Gritsenko, Chan, Norouzi, and
  Fleet]{ho2022videodiffusion}
Jonathan Ho, Tim Salimans, Alexey Gritsenko, William Chan, Mohammad Norouzi,
  and David~J. Fleet.
\newblock Video diffusion models.
\newblock \emph{arXiv preprint arXiv:2204.03458}, 2022.

\bibitem[Hoffmann et~al.(2022)Hoffmann, Borgeaud, Mensch, Buchatskaya, Cai,
  Rutherford, de~Las~Casas, Hendricks, Welbl, Clark, Hennigan, Noland,
  Millican, van~den Driessche, Damoc, Guy, Osindero, Simonyan, Elsen, Rae,
  Vinyals, and Sifre]{hoffman2022chinchilla}
Jordan Hoffmann, Sebastian Borgeaud, Arthur Mensch, Elena Buchatskaya, Trevor
  Cai, Eliza Rutherford, Diego de~Las~Casas, Lisa~Anne Hendricks, Johannes
  Welbl, Aidan Clark, Tom Hennigan, Eric Noland, Katie Millican, George van~den
  Driessche, Bogdan Damoc, Aurelia Guy, Simon Osindero, Karen Simonyan, Erich
  Elsen, Jack~W. Rae, Oriol Vinyals, and Laurent Sifre.
\newblock Training compute-optimal large language models.
\newblock \emph{arXiv preprint arXiv:2203.15556}, 2022.

\bibitem[Hoogeboom et~al.(2023)Hoogeboom, Heek, and
  Salimans]{hoogeboom2023simple}
Emiel Hoogeboom, Jonathan Heek, and Tim Salimans.
\newblock Simple diffusion: End-to-end diffusion for high resolution images.
\newblock In \emph{International Conference on Machine Learning (ICML)}, 2023.

\bibitem[Hu et~al.(2026)Hu, Qiu, Lu, Zhao, Li, Kim, Andreas, and He]{hu2026elf}
Keya Hu, Linlu Qiu, Yiyang Lu, Hanhong Zhao, Tianhong Li, Yoon Kim, Jacob
  Andreas, and Kaiming He.
\newblock Elf: Embedded language flows.
\newblock \emph{arXiv preprint arXiv:2605.10938}, 2026.

\bibitem[Jiang et~al.(2026)Jiang, Ren, Li, Wang, Sun, Yang, Liu, Zhang, Zheng,
  Zhang, Huang, Duan, and Zuo]{jiang2026textldm}
Jiaxiu Jiang, Jingjing Ren, Wenbo Li, Bo~Wang, Haoze Sun, Yijun Yang, Jianhui
  Liu, Yanbing Zhang, Shenghe Zheng, Yuan Zhang, Haoyang Huang, Nan Duan, and
  Wangmeng Zuo.
\newblock Textldm: Language modeling with continuous latent diffusion.
\newblock \emph{arXiv preprint arXiv:2605.07748}, 2026.

\bibitem[Karras et~al.(2022)Karras, Aittala, Aila, and Laine]{karras2022edm}
Tero Karras, Miika Aittala, Timo Aila, and Samuli Laine.
\newblock Elucidating the design space of diffusion-based generative models.
\newblock In \emph{Advances in Neural Information Processing Systems
  (NeurIPS)}, 2022.

\bibitem[Kong et~al.(2021)Kong, Ping, Huang, Zhao, and
  Catanzaro]{kong2020diffwave}
Zhifeng Kong, Wei Ping, Jiaji Huang, Kexin Zhao, and Bryan Catanzaro.
\newblock Diffwave: A versatile diffusion model for audio synthesis.
\newblock In \emph{International Conference on Learning Representations
  (ICLR)}, 2021.

\bibitem[Lai et~al.(2017)Lai, Xie, Liu, Yang, and Hovy]{lai2017race}
Guokun Lai, Qizhe Xie, Hanxiao Liu, Yiming Yang, and Eduard Hovy.
\newblock {RACE}: Large-scale {ReA}ding comprehension dataset from
  examinations.
\newblock In \emph{Proceedings of the 2017 Conference on Empirical Methods in
  Natural Language Processing}, 2017.

\bibitem[Li and He(2025)]{li2025jit}
Tianhong Li and Kaiming He.
\newblock Back to basics: Let denoising generative models denoise.
\newblock \emph{arXiv preprint arXiv:2511.13720}, 2025.

\bibitem[Li et~al.(2022)Li, Thickstun, Gulrajani, Liang, and
  Hashimoto]{li2022diffusionlm}
Xiang~Lisa Li, John Thickstun, Ishaan Gulrajani, Percy Liang, and Tatsunori~B.
  Hashimoto.
\newblock Diffusion-lm improves controllable text generation.
\newblock In \emph{Advances in Neural Information Processing Systems
  (NeurIPS)}, 2022.

\bibitem[Lin(2004)]{lin2004rouge}
Chin-Yew Lin.
\newblock {ROUGE}: A package for automatic evaluation of summaries.
\newblock In \emph{Text Summarization Branches Out}, pages 74--81, Barcelona,
  Spain, 2004. Association for Computational Linguistics.
\newblock URL \url{https://aclanthology.org/W04-1013/}.

\bibitem[Lipman et~al.(2023)Lipman, Chen, Ben-Hamu, Nickel, and
  Le]{lipman2023flow}
Yaron Lipman, Ricky T.~Q. Chen, Heli Ben-Hamu, Maximilian Nickel, and Matt Le.
\newblock Flow matching for generative modeling.
\newblock In \emph{International Conference on Learning Representations
  (ICLR)}, 2023.

\bibitem[Liu et~al.(2023)Liu, Gong, and Liu]{liu2023rectified}
Xingchao Liu, Chengyue Gong, and Qiang Liu.
\newblock Flow straight and fast: Learning to generate and transfer data with
  rectified flow.
\newblock In \emph{International Conference on Learning Representations
  (ICLR)}, 2023.

\bibitem[Lou et~al.(2024)Lou, Meng, and Ermon]{lou2024sedd}
Aaron Lou, Chenlin Meng, and Stefano Ermon.
\newblock Discrete diffusion modeling by estimating the ratios of the data
  distribution.
\newblock In \emph{Proceedings of the 41st International Conference on Machine
  Learning}, volume 235 of \emph{Proceedings of Machine Learning Research},
  pages 32819--32848, 2024.

\bibitem[Lovelace et~al.(2023)Lovelace, Kishore, Wan, Shekhtman, and
  Weinberger]{lovelace2023ld4lg}
Justin Lovelace, Varsha Kishore, Chao Wan, Eliot Shekhtman, and Kilian~Q.
  Weinberger.
\newblock Latent diffusion for language generation.
\newblock In \emph{Advances in Neural Information Processing Systems
  (NeurIPS)}, 2023.

\bibitem[Lu et~al.(2022)Lu, Zhou, Bao, Chen, Li, and Zhu]{lu2022dpmsolverpp}
Cheng Lu, Yuhao Zhou, Fan Bao, Jianfei Chen, Chongxuan Li, and Jun Zhu.
\newblock {DPM-Solver++}: Fast solver for guided sampling of diffusion
  probabilistic models.
\newblock \emph{arXiv preprint arXiv:2211.01095}, 2022.

\bibitem[Luo et~al.(2023)Luo, Tan, Huang, Li, and
  Zhao]{luo2023latentconsistency}
Simian Luo, Yiqin Tan, Longbo Huang, Jian Li, and Hang Zhao.
\newblock Latent consistency models: Synthesizing high-resolution images with
  few-step inference.
\newblock \emph{arXiv preprint arXiv:2310.04378}, 2023.

\bibitem[Meshchaninov et~al.(2025)Meshchaninov, Chimbulatov, Shabalin, Abramov,
  and Vetrov]{meshchaninov2025cosmos}
Viacheslav Meshchaninov, Egor Chimbulatov, Alexander Shabalin, Aleksandr
  Abramov, and Dmitry Vetrov.
\newblock Cosmos: Compressed and smooth latent space for text diffusion
  modeling.
\newblock In \emph{Advances in Neural Information Processing Systems
  (NeurIPS)}, 2025.

\bibitem[Mihaylov et~al.(2018)Mihaylov, Clark, Khot, and
  Sabharwal]{mihaylov2018openbookqa}
Todor Mihaylov, Peter Clark, Tushar Khot, and Ashish Sabharwal.
\newblock Can a suit of armor conduct electricity? {A} new dataset for open
  book question answering.
\newblock In \emph{Proceedings of the 2018 Conference on Empirical Methods in
  Natural Language Processing}, 2018.

\bibitem[Mostafazadeh et~al.(2016)Mostafazadeh, Chambers, He, Parikh, Batra,
  Vanderwende, Kohli, and Allen]{mostafazadeh2016storycloze}
Nasrin Mostafazadeh, Nathanael Chambers, Xiaodong He, Devi Parikh, Dhruv Batra,
  Lucy Vanderwende, Pushmeet Kohli, and James Allen.
\newblock A corpus and cloze evaluation for deeper understanding of commonsense
  stories.
\newblock In \emph{Proceedings of the 2016 Conference of the North American
  Chapter of the Association for Computational Linguistics}, 2016.

\bibitem[Narayan et~al.(2018)Narayan, Cohen, and Lapata]{narayan2018xsum}
Shashi Narayan, Shay~B. Cohen, and Mirella Lapata.
\newblock Don't give me the details, just the summary! topic-aware
  convolutional neural networks for extreme summarization.
\newblock In \emph{Proceedings of the 2018 Conference on Empirical Methods in
  Natural Language Processing}, pages 1797--1807, 2018.

\bibitem[Nie et~al.(2025)Nie, Zhu, You, Zhang, Ou, Hu, Zhou, Lin, Wen, and
  Li]{nie2025llada}
Shen Nie, Fengqi Zhu, Zebin You, Xiaolu Zhang, Jingyang Ou, Jun Hu, Jun Zhou,
  Yankai Lin, Ji-Rong Wen, and Chongxuan Li.
\newblock Large language diffusion models.
\newblock \emph{arXiv preprint arXiv:2502.09992}, 2025.

\bibitem[OpenAI et~al.(2023)OpenAI, Achiam, Adler, Agarwal, Ahmad, Akkaya,
  et~al.]{openai2024gpt4}
OpenAI, Josh Achiam, Steven Adler, Sandhini Agarwal, Lama Ahmad, Ilge Akkaya,
  et~al.
\newblock {GPT-4} technical report.
\newblock \emph{arXiv preprint arXiv:2303.08774}, 2023.

\bibitem[Pillutla et~al.(2021)Pillutla, Swayamditta, Zettlemoyer, and
  Harchaoui]{pillutla2021mauve}
Krishna Pillutla, Swabha Swayamditta, Luke Zettlemoyer, and Zaid Harchaoui.
\newblock {MAUVE}: Measuring the gap between neural text and human text using
  divergence frontiers.
\newblock In \emph{Advances in Neural Information Processing Systems}, 2021.

\bibitem[Radford et~al.(2019)Radford, Wu, Child, Luan, Amodei, and
  Sutskever]{radford2019language}
Alec Radford, Jeffrey Wu, Rewon Child, David Luan, Dario Amodei, and Ilya
  Sutskever.
\newblock Language models are unsupervised multitask learners.
\newblock OpenAI technical report, 2019.

\bibitem[Rajpurkar et~al.(2016)Rajpurkar, Zhang, Lopyrev, and
  Liang]{rajpurkar2016squad}
Pranav Rajpurkar, Jian Zhang, Konstantin Lopyrev, and Percy Liang.
\newblock {SQ}u{AD}: 100,000+ questions for machine comprehension of text.
\newblock In \emph{Proceedings of the 2016 Conference on Empirical Methods in
  Natural Language Processing}, 2016.

\bibitem[Rombach et~al.(2022)Rombach, Blattmann, Lorenz, Esser, and
  Ommer]{rombach2022ldm}
Robin Rombach, Andreas Blattmann, Dominik Lorenz, Patrick Esser, and Bj{\"o}rn
  Ommer.
\newblock High-resolution image synthesis with latent diffusion models.
\newblock In \emph{IEEE/CVF Conference on Computer Vision and Pattern
  Recognition (CVPR)}, 2022.

\bibitem[Sahoo et~al.(2024)Sahoo, Arriola, Schiff, Gokaslan, Marroquin, Chiu,
  Rush, and Kuleshov]{sahoo2024mdlm}
Subham~Sekhar Sahoo, Marianne Arriola, Yair Schiff, Aaron Gokaslan, Edgar
  Marroquin, Justin~T. Chiu, Alexander~M. Rush, and Volodymyr Kuleshov.
\newblock Simple and effective masked diffusion language models.
\newblock In \emph{Advances in Neural Information Processing Systems
  (NeurIPS)}, 2024.

\bibitem[Sahoo et~al.(2025)Sahoo, Deschenaux, Gokaslan, Wang, Chiu, and
  Kuleshov]{sahoo2025duo}
Subham~Sekhar Sahoo, Justin Deschenaux, Aaron Gokaslan, Guanghan Wang,
  Justin~T. Chiu, and Volodymyr Kuleshov.
\newblock The diffusion duality.
\newblock In \emph{International Conference on Machine Learning (ICML)}, 2025.

\bibitem[Sakaguchi et~al.(2020)Sakaguchi, Le~Bras, Bhagavatula, and
  Choi]{sakaguchi2020winogrande}
Keisuke Sakaguchi, Ronan Le~Bras, Chandra Bhagavatula, and Yejin Choi.
\newblock {WinoGrande}: An adversarial winograd schema challenge at scale.
\newblock In \emph{Proceedings of the 34th AAAI Conference on Artificial
  Intelligence}, 2020.

\bibitem[Sap et~al.(2019)Sap, Rashkin, Chen, Le~Bras, and Choi]{sap2019siqa}
Maarten Sap, Hannah Rashkin, Derek Chen, Ronan Le~Bras, and Yejin Choi.
\newblock Social {IQa}: Commonsense reasoning about social interactions.
\newblock In \emph{Proceedings of the 2019 Conference on Empirical Methods in
  Natural Language Processing}, 2019.

\bibitem[Shabalin et~al.(2025)Shabalin, Meshchaninov, Chimbulatov, Lapikov,
  Kim, Bartosh, Molchanov, Markov, and Vetrov]{shabalin2024tencdm}
Alexander Shabalin, Viacheslav Meshchaninov, Egor Chimbulatov, Vladislav
  Lapikov, Roman Kim, Grigory Bartosh, Dmitry Molchanov, Sergey Markov, and
  Dmitry Vetrov.
\newblock Tencdm: Understanding the properties of the diffusion model in the
  space of language model encodings.
\newblock \emph{Proceedings of the AAAI Conference on Artificial Intelligence},
  39\penalty0 (23):\penalty0 25110--25118, 2025.

\bibitem[Shazeer(2020)]{shazeer2020glu}
Noam Shazeer.
\newblock {GLU} variants improve transformer.
\newblock \emph{arXiv preprint arXiv:2002.05202}, 2020.

\bibitem[Shen et~al.(2026)Shen, Zhao, He, and Lin]{shen2026codar}
Junzhe Shen, Jieru Zhao, Ziwei He, and Zhouhan Lin.
\newblock Codar: Continuous diffusion language models are more powerful than
  you think.
\newblock \emph{arXiv preprint arXiv:2603.02547}, 2026.

\bibitem[Song et~al.(2021)Song, Sohl-Dickstein, Kingma, Kumar, Ermon, and
  Poole]{song2021scoresde}
Yang Song, Jascha Sohl-Dickstein, Diederik~P. Kingma, Abhishek Kumar, Stefano
  Ermon, and Ben Poole.
\newblock Score-based generative modeling through stochastic differential
  equations.
\newblock In \emph{International Conference on Learning Representations
  (ICLR)}, 2021.

\bibitem[Song et~al.(2023)Song, Dhariwal, Chen, and
  Sutskever]{song2023consistency}
Yang Song, Prafulla Dhariwal, Mark Chen, and Ilya Sutskever.
\newblock Consistency models.
\newblock In \emph{International Conference on Machine Learning (ICML)}, 2023.

\bibitem[Song et~al.(2025)Song, Zhang, Luo, Gao, Xia, Luo, Li, Yang, Yu, Qu,
  Fu, Su, Zhang, Huang, Wang, Yan, Jia, Liu, Ma, Zhang, Wu, and
  Zhou]{seedteam2025seeddiffusion}
Yuxuan Song, Zheng Zhang, Cheng Luo, Pengyang Gao, Fan Xia, Hao Luo, Zheng Li,
  Yuehang Yang, Hongli Yu, Xingwei Qu, Yuwei Fu, Jing Su, Ge~Zhang, Wenhao
  Huang, Mingxuan Wang, Lin Yan, Xiaoying Jia, Jingjing Liu, Wei-Ying Ma,
  Ya-Qin Zhang, Yonghui Wu, and Hao Zhou.
\newblock Seed diffusion: A large-scale diffusion language model with
  high-speed inference.
\newblock \emph{arXiv preprint arXiv:2508.02193}, 2025.

\bibitem[Su et~al.(2021)Su, Lu, Pan, Murtadha, Wen, and Liu]{su2021roformer}
Jianlin Su, Yu~Lu, Shengfeng Pan, Ahmed Murtadha, Bo~Wen, and Yunfeng Liu.
\newblock {RoFormer}: Enhanced transformer with rotary position embedding.
\newblock \emph{arXiv preprint arXiv:2104.09864}, 2021.

\bibitem[Touvron et~al.(2023)Touvron, Lavril, Izacard, Martinet,
  et~al.]{touvron2023llama}
Hugo Touvron, Thibaut Lavril, Gautier Izacard, Xavier Martinet, et~al.
\newblock Llama: Open and efficient foundation language models.
\newblock \emph{arXiv preprint arXiv:2302.13971}, 2023.

\bibitem[Vaswani et~al.(2017)Vaswani, Shazeer, Parmar, Uszkoreit, Jones, Gomez,
  Kaiser, and Polosukhin]{vaswani2017attention}
Ashish Vaswani, Noam Shazeer, Niki Parmar, Jakob Uszkoreit, Llion Jones,
  Aidan~N. Gomez, Lukasz Kaiser, and Illia Polosukhin.
\newblock Attention is all you need.
\newblock In \emph{Advances in Neural Information Processing Systems},
  volume~30, 2017.

\bibitem[Wang et~al.(2024)Wang, Zhang, Luo, Sun, Cui, Wang, Zhang, Wang, Li,
  Yu, et~al.]{wang2024emu3}
Xinlong Wang, Xiaosong Zhang, Zhengxiong Luo, Quan Sun, Yufeng Cui, Jinsheng
  Wang, Fan Zhang, Yueze Wang, Zhen Li, Qiying Yu, et~al.
\newblock Emu3: Next-token prediction is all you need.
\newblock \emph{arXiv preprint arXiv:2409.18869}, 2024.

\bibitem[Yang et~al.(2025)Yang, Li, Yang, Zhang, et~al.]{yang2024qwen3}
An~Yang, Anfeng Li, Baosong Yang, Beichen Zhang, et~al.
\newblock Qwen3 technical report.
\newblock \emph{arXiv preprint arXiv:2505.09388}, 2025.

\bibitem[Yang et~al.(2026)Yang, Guo, Zhang, Sahoo, Chen, Vahdat, Mardani, and
  Thickstun]{yang2026replaid}
Zhihan Yang, Wei Guo, Shuibai Zhang, Subham~Sekhar Sahoo, Yongxin Chen, Arash
  Vahdat, Morteza Mardani, and John Thickstun.
\newblock Continuous diffusion scales competitively with discrete diffusion for
  language.
\newblock \emph{arXiv preprint arXiv:2605.18530}, 2026.

\bibitem[Ye et~al.(2025)Ye, Xie, Zheng, Gao, Wu, Jiang, Li, and
  Kong]{ye2025dream}
Jiacheng Ye, Zhihui Xie, Lin Zheng, Jiahui Gao, Zirui Wu, Xin Jiang, Zhenguo
  Li, and Lingpeng Kong.
\newblock Dream 7b: Diffusion large language models.
\newblock \emph{arXiv preprint arXiv:2508.15487}, 2025.

\bibitem[Yuan et~al.(2024)Yuan, Yuan, Tan, Huang, and
  Huang]{yuan2024seqdiffuseq}
Hongyi Yuan, Zheng Yuan, Chuanqi Tan, Fei Huang, and Songfang Huang.
\newblock Text diffusion model with encoder-decoder transformers for
  sequence-to-sequence generation.
\newblock In \emph{Proceedings of the 2024 Conference of the North American
  Chapter of the Association for Computational Linguistics: Human Language
  Technologies (Volume 1: Long Papers)}, pages 22--39, Mexico City, Mexico,
  2024. Association for Computational Linguistics.
\newblock \doi{10.18653/v1/2024.naacl-long.2}.

\bibitem[Zellers et~al.(2019)Zellers, Holtzman, Bisk, Farhadi, and
  Choi]{zellers2019hellaswag}
Rowan Zellers, Ari Holtzman, Yonatan Bisk, Ali Farhadi, and Yejin Choi.
\newblock {HellaSwag}: Can a machine really finish your sentence?
\newblock In \emph{Proceedings of the 57th Annual Meeting of the Association
  for Computational Linguistics}, 2019.

\bibitem[Zhang and Sennrich(2019)]{zhang2019rmsnorm}
Biao Zhang and Rico Sennrich.
\newblock Root mean square layer normalization.
\newblock In \emph{Advances in Neural Information Processing Systems},
  volume~32, 2019.

\bibitem[Zhao et~al.(2025)Zhao, Gupta, Zheng, and Grover]{zhao2025d1}
Siyan Zhao, Devaansh Gupta, Qinqing Zheng, and Aditya Grover.
\newblock d1: Scaling reasoning in diffusion large language models via
  reinforcement learning.
\newblock \emph{arXiv preprint arXiv:2504.12216}, 2025.

\end{thebibliography}

\clearpage
\beginappendix
\addcontentsline{toc}{section}{Appendix}
\addtocontents{toc}{\protect\setcounter{tocdepth}{-1}}
\section{Additional Method and Experimental Details}
\label{app:details}

\makeatletter
\setlength{\@fptop}{0pt}
\setlength{\@fpsep}{12pt}
\setlength{\@fpbot}{0pt plus 1fil}
\makeatother

\definecolor{auroraPurple}{RGB}{91,0,153}
\definecolor{auroraLavender}{RGB}{249,246,252}
\setlength{\fboxsep}{8pt}
\setlength{\fboxrule}{0.55pt}

This appendix supplements the main text with implementation details, evaluation protocols, complete controlled-study results, and qualitative examples.
It begins with the remaining method specifications, followed by the datasets, model configurations, training procedures, and evaluation settings, and concludes with the full quantitative results and generated samples.

\subsection{Method and Implementation Details}
\label{app:method-details}

\subsubsection{Autoencoder Training-Time Masking}
\label{app:ae-masking}

We apply token-embedding dropout and latent dropout during autoencoder training~\citep{meshchaninov2025cosmos}. 
This subsection specifies these two masking operators used in our implementation.
Using the trainable embedding matrix $E\in\mathbb{R}^{|V|\times d}$ defined in
Sec.~\ref{sec:ae}, token-embedding dropout independently zeros the input vector at each valid token position. 
For $j\in\{1,\ldots,L\}$, let $m_j\sim\mathrm{Bernoulli}(p_x)$; the masked encoder input is
\begin{equation}
  \widetilde E[w_j] = (1-m_j)E[w_j],
  \qquad p_x=0.3 .
  \label{eq:app-input-masking}
\end{equation}
After encoding, latent dropout independently masks each coordinate of
$z_{\mathrm{enc}}\in\mathbb{R}^{N\times D}$ before decoding:
\begin{equation}
  \widetilde z_{\mathrm{enc}} = z_{\mathrm{enc}} \odot r,\qquad
  r_{ik}\sim\mathrm{Bernoulli}(1-p_z),
  \quad i\in\{1,\ldots,N\},\ k\in\{1,\ldots,D\},
  \qquad p_z=0.6 ,
  \label{eq:app-latent-masking}
\end{equation}
with no inverted-dropout rescaling.

\subsubsection{Loss-Space Conversion for Clean-Latent Prediction}
\label{app:target-loss}

Under the linear interpolant used in flow matching~\citep{lipman2023flow,liu2023rectified}, Eq.~\eqref{eq:fm-interpolant} uses $t$ as the interpolation coefficient.
To match the notation of the controlled study, we denote this coefficient by $\sigma$.
Setting $t=\sigma$ and $x_0=z$, where $z$ denotes the clean latent, gives the noisy latent and its path velocity as
\begin{equation}
  x_\sigma=(1-\sigma)z+\sigma\varepsilon,
  \qquad
  v=\varepsilon-z,
  \qquad \varepsilon\sim\mathcal{N}(0,I).
  \label{eq:app-target-loss-path}
\end{equation}
If the model predicts the clean latent $\hat z_\theta(x_\sigma,\sigma)$, the corresponding velocity prediction is
\begin{equation}
  \hat v_\theta(x_\sigma,\sigma)
  = \frac{x_\sigma-\hat z_\theta(x_\sigma,\sigma)}{\sigma},
  \qquad
  \|\hat v_\theta-v\|_2^2
  = \frac{1}{\sigma^2}
  \|\hat z_\theta(x_\sigma,\sigma)-z\|_2^2 .
  \label{eq:app-v-loss-reweight}
\end{equation}
Therefore, evaluating a clean-latent prediction in velocity space multiplies its clean-latent error by $1/\sigma^2$, assigning greater weight to samples at lower noise levels. 
This is the loss-space effect examined in Sec.~\ref{sec:study-target}.

\subsubsection{High-Noise Mass of the Schedule Families}
\label{app:tand-schedule}

The tan-$d$ mapping used in the main text follows a shifted cosine parameterization~\citep{hoogeboom2023simple}.
Here $\sigma=t$ in the linear interpolant. The implementation samples $\tau\sim\mathcal U(10^{-3},1)$ and clips the resulting noise level to $[10^{-5},1-10^{-5}]$.
To compare different noise-level distributions, Sec.~\ref{sec:study-schedule} uses the high-noise mass $m_{\sigma_\star}=\Pr(\sigma>\sigma_\star)$ at a threshold $\sigma_\star\in(0,1)$. 
For the idealized tan-$d$ mapping $\sigma(\tau;d)$ defined in the main text, before the implementation cutoffs, $\tau\sim\mathcal U(0,1)$ gives
\begin{equation}
  m_{\sigma_\star}
  =\Pr[\sigma(\tau;d)>\sigma_\star]
  =
  1-\frac{2}{\pi}\arctan\!\left(
  \frac{\sigma_\star}{d(1-\sigma_\star)}\right).
  \label{eq:app-tand-mass}
\end{equation}
For logit-normal sampling, $\ell\sim\mathcal N(\mu,s^2)$ and
$\sigma=\operatorname{sigmoid}(\ell)$. Since the sigmoid mapping is monotone,
the corresponding high-noise mass is
\begin{equation}
m_{\sigma_\star}
=
\Pr[\sigma>\sigma_\star]
=
1-\Phi\!\left(
\frac{\operatorname{logit}(\sigma_\star)-\mu}{s}
\right),
\end{equation}
where $\Phi$ denotes the standard normal cumulative distribution function.
Setting $\sigma_\star=0.7$ gives the same $m_{0.7}$ coordinate used to compare
the tan-$d$ and logit-normal sweeps in Sec.~\ref{sec:study-schedule}.

\subsubsection{Self-Trajectory Consistency Implementation}
\label{app:trajectory-consistency}

During STC training, we obtain $t'=\operatorname{Prev}_{S}(t)$ by taking a step of size $1/S$ toward lower noise in the tan-$d$ schedule coordinate.
For the tan-$d$ mapping in Eq.~\eqref{eq:study-tand-map},
\begin{equation}
  \tau=\frac{2}{\pi}\arctan\!\left(\frac{t}{d(1-t)}\right),
  \qquad t'=\sigma\!\left(\max\{\tau-1/S,10^{-3}\};d\right).
  \label{eq:app-stc-prev}
\end{equation}
The resulting $t'$ is clipped to $[10^{-5},1-10^{-5}]$.
The state update and consistency loss follow Eqs.~\eqref{eq:consistency-neighbor} and~\eqref{eq:consistency-loss}.
For conditional training, visible prompt positions are restored to their clean latents after the update, using the same conditioning-dropout mask.
When self-conditioning is enabled, the EMA input is the detached current prediction, including updates without a draft estimate for the current model.

The consistency weight is increased linearly from zero to its final value
$\lambda_{\mathrm{ct}}$ over the first $T_{\mathrm{warm}}$ optimization steps.
Unless self-trajectory consistency is explicitly disabled, the reported runs use
\begin{equation}
  \lambda_{\mathrm{ct}}=1.0,\qquad S=16.
  \label{eq:app-ct-config}
\end{equation}
We set $T_{\mathrm{warm}}=10{,}000$ optimization steps for AURORA-LM-S
and $T_{\mathrm{warm}}=5{,}000$ optimization steps for AURORA-LM-L.

\subsection{Experimental Protocol and System Configuration}
\label{app:protocol-details}

This section specifies the data preprocessing, model architecture, training configuration, and generation and evaluation protocols used throughout the experiments. 
It also documents the inference sweep used to select the system-level OpenWebText configuration, together with the baseline-reproduction and public-benchmark evaluation procedures.

\subsubsection{Training-Data Preprocessing}
\label{app:data-preprocessing}

For both the controlled design analysis and the system-level comparison, documents are drawn from OpenWebText~\citep{gokaslan2019openwebtext}.
The controlled analysis uses the Qwen3 tokenizer with a maximum sequence length of 128 tokens (OWT128), whereas the system-level comparison uses the GPT-2 BPE tokenizer and packs documents into sequences of 1,024 tokens. For a fair system-level comparison, we follow the OpenWebText data split and preprocessing protocol used by Duo~\citep{sahoo2025duo}, including fixed-length wrap packing. 
Each sequence begins and ends with an EOS token, and adjacent documents within a packed sequence are separated by EOS.

For AURORA-LM-L, we construct a 294.7B-token training mixture from diverse English-language resources, including public web, books, academic and mathematical text, as well as synthetically generated text. After source-level weighting and packing, the mixture comprises 161.3B web tokens (54.7\%), 50.3B tokens from books and encyclopedia (17.1\%), and 83.1B tokens of academic, mathematical, and knowledge text (28.2\%). These counts describe the constructed corpus rather than the total token exposures accumulated over optimization. The model uses Qwen3 tokenization, and the resulting documents are packed into 512-token sequences.

\subsubsection{Model Architecture}

\begin{table}[!ht]
  \centering
  \small
  \setlength{\tabcolsep}{4pt}
  \caption{\textbf{Architectures of AURORA-LM-S and AURORA-LM-L.} AURORA-LM-S is used
  for the system-level comparisons, whereas AURORA-LM-L is used for the
  scaling evaluation. Model-size labels refer to the block-causal denoiser and
  exclude the autoencoder.}
  \label{tab:app-aurora-model}
  \begin{tabularx}{\linewidth}{p{0.28\linewidth}XX}
    \toprule
    \textbf{Component} & \textbf{AURORA-LM-S (130M)} & \textbf{AURORA-LM-L (1B)} \\
    \midrule
    \multicolumn{3}{l}{\emph{Query-based Encoder-Decoder}} \\
    Encoder layers / decoder layers & 6 / 6 & 6 / 6 \\
    Hidden size / heads / head dim & 768 / 12 / 64 & 1024 / 16 / 64 \\
    Feed-forward intermediate & 3072 & 4096 \\
    Latent channel $D$ & 768 & 1024 \\
    Latent length $M$ & 1024 ($c{=}1$) & 512 ($c{=}1$) \\
    Token-embedding initialization & GPT-2 small & Qwen3-0.6B \\
    Encoder parameters & 49.6M & 88.1M \\
    Decoder parameters & 88.2M, including tied output projection & 243.7M, including tied output projection \\
    \midrule
    \multicolumn{3}{l}{\emph{Block-causal denoiser}} \\
    Layers & 12 & 24 \\
    Hidden size / heads / head dim & 768 / 12 / 64 & 1536 / 12 / 128 \\
    Feed-forward intermediate & 3072 & 6144 \\
    Noisy-input bottleneck $D_b$ & 128 & 128 \\
    Block size $Q$ & 16 & 16 \\
    Positional encoding & RoPE ($\theta{=}10{,}000$, max\_pos 1024) & RoPE ($\theta{=}10{,}000$, max\_pos 512) \\
    Attention mode & Two-stream: clean prefix to noised current block & Two-stream: clean prefix to noised current block \\
    Denoiser parameters & 130M & 1.01B \\
    \midrule
    Total system parameters & 267.8M & $\approx$1.34B \\
    \bottomrule
  \end{tabularx}
\end{table}

The token-embedding tables are initialized from the listed pretrained models and updated during autoencoder training.
The trained autoencoder, including its tied output embedding, is then frozen before denoiser training.

The block-causal denoiser is implemented as a Transformer~\citep{vaswani2017attention} using RMSNorm~\citep{zhang2019rmsnorm}, SwiGLU feed-forward layers~\citep{shazeer2020glu}, RoPE positional encoding~\citep{su2021roformer}, and per-head QK normalization~\citep{henry2020qknorm}.
Timestep conditioning uses sinusoidal embeddings~\citep{ho2020ddpm} projected through a two-layer MLP and injected additively into each Transformer layer.

\clearpage

\subsubsection{Training Configuration}

\begin{table}[!ht]
  \centering
  \small
  \setlength{\tabcolsep}{4pt}
  \caption{\textbf{Training configurations for AURORA-LM-S and AURORA-LM-L
in Table~\ref{tab:app-aurora-model}.}
The underlying objectives are defined in \S\ref{sec:method} and
Appendix~\ref{app:method-details}.
Optimization steps, warmup durations, and EMA frequencies refer to parameter
updates; global batch sizes are reported after gradient accumulation.
Training compute is estimated using $3ND$ accounting, with one
multiply--accumulate counted as one operation.}
  \label{tab:app-aurora-training}
  \begin{tabularx}{\linewidth}{p{0.28\linewidth}XX}
    \toprule
    \textbf{Setting} & \textbf{AURORA-LM-S (130M)} & \textbf{AURORA-LM-L (1B)} \\
    \midrule
    \multicolumn{3}{l}{\emph{Shared (Autoencoder + Denoiser)}} \\
    Optimizer & AdamW, $\beta{=}(0.9,0.98)$ & AdamW, $\beta{=}(0.9,0.98)$ \\
    Gradient clipping & 1.0 & 1.0 \\
    Mixed precision & bf16 & bf16 \\
    Random seed & 42 & 42 \\
    \midrule
    \multicolumn{3}{l}{\emph{Autoencoder training}} \\
    Optimization steps & 100{,}000 & 150{,}000 \\
Global batch size & 1024 & 2048 \\
    Weight decay & $10^{-2}$ & $10^{-2}$ \\
    Learning rate & $4\times10^{-4}$ & $2\times10^{-4}$ \\
    Token-embedding dropout $p_x$ & 0.3 & 0.3 \\
    Latent feature dropout $p_z$ & 0.6 & 0.6 \\
    Training compute & $\approx 43$ EFLOPs & $\approx 157$ EFLOPs \\
    \midrule
    \multicolumn{3}{l}{\emph{Denoiser training}} \\
    Optimization steps & 250{,}000 (OWT); 300{,}000 (XSum) & 590{,}000 \\
Global batch size & 1024 (OWT); 128 (XSum) & 1024 \\
    $\varepsilon$ & $10^{-6}$ & $10^{-6}$ \\
    Learning rate & $4\times10^{-4}$ & Peak $2\times10^{-4}$, minimum $1\times10^{-6}$ \\
    Weight decay & $10^{-5}$ & $10^{-5}$ \\
    Warmup & 1000 steps & 1000 steps \\
    EMA & 0.9999, every optimization step
    & 0.9999, every optimization step \\
    Parameterization & Flow matching, $x_\sigma=(1-\sigma)z+\sigma\varepsilon$ & Flow matching, $x_\sigma=(1-\sigma)z+\sigma\varepsilon$ \\
    Prediction target & $x_0$ (clean latent) & $x_0$ (clean latent) \\
    Loss space & $x_0$-space & $x_0$-space \\
    Time sampling & tan-$d$, shift~$7.0$ & tan-$d$, shift~$7.0$ \\
    Self-conditioning & $p_{\mathrm{sc}}{=}0.6$ & $p_{\mathrm{sc}}{=}0.5$ \\
    Self-trajectory consistency & $\lambda_{\mathrm{ct}}{=}1.0$, $S{=}16$ & $\lambda_{\mathrm{ct}}{=}1.0$, $S{=}16$ \\
    Consistency warmup & 10{,}000 steps & 5{,}000 steps \\
    Latent standardization & Per-channel, held-out encoder statistics & Per-channel, held-out encoder statistics \\
    Instruction fine-tuning & None & None \\
    Training compute & $\approx 155$ EFLOPs (OWT); $\approx 23$ EFLOPs (XSUM) & $\approx 1{,}406$ EFLOPs \\
    \bottomrule
  \end{tabularx}
\end{table}

\clearpage

\subsubsection{Controlled OWT128 Studies}
\label{app:owt128-protocol}

Table~\ref{tab:app-owt128-reference} summarizes the default configuration for the controlled OWT128 studies in Sec.~\ref{sec:study}.

\begin{table}[htbp]
\centering
\small
\caption{\textbf{Default configuration for the controlled OWT128 studies.} Unless otherwise specified, settings follow this configuration. Sweep-specific widths, schedules, block sizes, and sampling budgets are given with the full results in Appendix~\ref{app:design-study-diagnostics}.}
\label{tab:app-owt128-reference}
\begin{tabular}{ll}
\toprule
\textbf{Setting} & \textbf{Default value} \\
\midrule
\multicolumn{2}{l}{\emph{Architecture and latent configuration}} \\
Maximum token sequence length & 128 \\
Encoder layers / decoder layers & 6 / 6 \\
AE hidden / feed-forward width & 1024 / 4096 \\
Token-embedding initialization & Qwen3-0.6B \\
Denoiser layers & 12 \\
Denoiser hidden / heads / head dim & 1024 / 16 / 64 \\
Denoiser feed-forward width & 4096 \\
Latent width $D$ / retention ratio $c$ & 1024 / 1 \\
Noisy-input width $D_b$ / block size $Q$ & 128 / 16 \\
\midrule
\multicolumn{2}{l}{\emph{Autoencoder and denoiser training}} \\
Optimizer (both stages) & AdamW, $\beta=(0.9,0.98)$ \\
Optimization steps (AE / denoiser) & 50{,}000 / 50{,}000 \\
Global batch size (both stages) & 1024 \\
Learning rate (both stages) & $4\times10^{-4}$ \\
Linear warmup (both stages) & 1000 optimization steps \\
Learning-rate schedule & Constant after warmup \\
Weight decay (AE / denoiser) & $10^{-2}$ / $10^{-5}$ \\
Gradient clipping (both stages) & 1.0 \\
Noise allocation & tan-$d$, $d=5$ \\
Prediction target / loss space & $x_0$ / $x_0$ \\
\midrule
\multicolumn{2}{l}{\emph{Evaluation}} \\
Evaluation sampler & 32-step ODE, no guidance \\
Generation seeds / samples per seed & 1--5 / 1,000 \\
\bottomrule
\end{tabular}
\end{table}

\FloatBarrier

\subsubsection{Generation and Evaluation Protocols}
\label{app:generation-protocols}

Controlled design studies (Sec.~\ref{sec:study}) use 32-step deterministic ODE sampling.
The system-level OpenWebText comparison (Sec.~\ref{sec:scale-uncond}) uses 64-step SDE-DPM++ sampling with self-conditioning classifier-free guidance (SC-CFG) at $w_{\mathrm{sc}}=4$.
The XSum conditional comparison (Sec.~\ref{sec:scale-cond}) uses 32-step ODE sampling with classifier-free guidance at scale $w=2.0$.
In both prefix-conditioned settings, CFG is applied only in the high-noise window $\sigma\in[0.9,1.0]$.

For prefix-conditioned generation (XSum and the public benchmark), we first
encode the prompt into a clean latent prefix and use its complete blocks as
causal context for subsequent blockwise generation, following
\citet{guo2026cola}. If the prefix terminates within a block, our
implementation uses clean repainting: the known prefix positions in this
boundary block are fixed to their encoder-produced latents throughout
denoising, while the remaining positions are generated.

The public benchmark uses 16 denoising steps, guidance scale $3.0$ applied in
the high-noise window $\sigma\in[0.9,1.0]$, and at most 32 newly generated
tokens. AURORA-LM-L benchmark evaluation uses a fixed base sampling seed of
$256$.

Cola-DLM is evaluated using its publicly released checkpoint and inference
implementation~\citep{guo2026cola}. We retain its release defaults: a 16-step
Euler sampler, CFG scale $7.0$, and per-sample inference-noise base seed $66$.
Both systems are evaluated on the same benchmark instances using the shared
prompt, truncation, and answer-matching protocol in
Sec.~\ref{app:benchmark-protocol}; sampling configurations remain
model-specific. We do not retrain Cola-DLM, so its training configuration is
not treated as matched to that of AURORA-LM.

\subsubsection{System-Level Inference-Protocol Selection}
\label{app:research-inference-sweep}

Table~\ref{tab:app-research-inference-sweep} reports the complete inference-protocol sweep underlying the system-level OpenWebText comparison.
At the fixed 250k checkpoint, we vary the denoising-step count, solver, and SC-CFG scale. 
Each configuration generates 1,000 samples for each of five evaluation seeds, and entries report the mean $\pm$ standard deviation.
The corresponding architecture and generation configurations for the
system-level baselines are listed in Table~\ref{tab:model_arch}.

Within a fixed solver and step budget, increasing SC-CFG generally lowers Gen-PPL until the improvement saturates. 
Its effect on MAUVE is non-monotonic and depends on both the solver and step budget.
We select the 64-step SDE-DPM++ configuration with $w_{\mathrm{sc}}=4$, which attains the lowest mean Gen-PPL in the sweep while retaining a MAUVE of $0.890$.

\begin{table}[htbp]
  \centering
  \scriptsize
  \renewcommand{\arraystretch}{1.04}
  \setlength{\tabcolsep}{2.3pt}
  \caption{\textbf{Complete inference-protocol sweep for AURORA-LM-S at the 250k
  checkpoint under the OpenWebText free-generation protocol.} Results are
  grouped by denoising-step count so that the two solvers can be compared at
  the same SC-CFG scale. Each entry reports the mean $\pm$ standard deviation
  over evaluation seeds 1--5, with 1,000 samples per seed. The shaded cells
  mark the selected system-level configuration.}
  \label{tab:app-research-inference-sweep}
  \begin{tabular}{@{}c ccc ccc@{}}
    \toprule
    \multirow{2}{*}{\textbf{SC-CFG}} &
    \multicolumn{3}{c}{\textbf{Euler}} &
    \multicolumn{3}{c}{\textbf{SDE-DPM++}} \\
    \cmidrule(lr){2-4}\cmidrule(l){5-7}
    & \textbf{Gen-PPL $\downarrow$} & \textbf{Entropy $\uparrow$} & \textbf{MAUVE $\uparrow$}
    & \textbf{Gen-PPL $\downarrow$} & \textbf{Entropy $\uparrow$} & \textbf{MAUVE $\uparrow$} \\
    \midrule
    \multicolumn{7}{l}{\emph{16 denoising steps}} \\
    1 & $72.72\pm0.44$ & $5.385\pm0.002$ & $0.8854\pm0.0240$
      & $98.36\pm1.36$ & $5.449\pm0.009$ & $0.8481\pm0.0271$ \\
    2 & $60.57\pm0.35$ & $5.365\pm0.001$ & $0.9036\pm0.0156$
      & $70.11\pm0.24$ & $5.412\pm0.007$ & $0.8888\pm0.0212$ \\
    3 & $57.27\pm0.12$ & $5.368\pm0.003$ & $0.9083\pm0.0114$
      & $61.43\pm0.29$ & $5.396\pm0.003$ & $0.9214\pm0.0211$ \\
    4 & $54.72\pm0.60$ & $5.367\pm0.007$ & $0.9199\pm0.0092$
      & $54.86\pm0.44$ & $5.361\pm0.004$ & $0.9199\pm0.0120$ \\
    5 & $54.92\pm0.47$ & $5.374\pm0.004$ & $0.9149\pm0.0184$
      & $54.67\pm0.52$ & $5.380\pm0.005$ & $0.8952\pm0.0215$ \\
    \midrule
    \multicolumn{7}{l}{\emph{32 denoising steps}} \\
    1 & $50.50\pm0.33$ & $5.323\pm0.005$ & $0.8930\pm0.0228$
      & $62.01\pm0.23$ & $5.406\pm0.002$ & $0.9232\pm0.0147$ \\
    2 & $41.58\pm0.42$ & $5.309\pm0.006$ & $0.9271\pm0.0072$
      & $41.53\pm0.53$ & $5.347\pm0.009$ & $0.9482\pm0.0143$ \\
    3 & $38.93\pm0.26$ & $5.314\pm0.004$ & $0.9440\pm0.0120$
      & $35.44\pm0.50$ & $5.310\pm0.007$ & $\mathbf{0.9516\pm0.0055}$ \\
    4 & $37.65\pm0.41$ & $5.316\pm0.004$ & $0.9344\pm0.0178$
      & $31.63\pm0.33$ & $5.292\pm0.004$ & $0.9250\pm0.0177$ \\
    5 & $37.09\pm0.19$ & $5.312\pm0.006$ & $0.9225\pm0.0103$
      & $31.95\pm0.13$ & $5.311\pm0.005$ & $0.9205\pm0.0173$ \\
    \midrule
    \multicolumn{7}{l}{\emph{64 denoising steps}} \\
    1 & $42.55\pm0.22$ & $5.326\pm0.002$ & $0.9258\pm0.0191$
      & $45.92\pm0.38$ & $5.391\pm0.005$ & $0.9407\pm0.0131$ \\
    2 & $34.01\pm0.12$ & $5.294\pm0.004$ & $0.9290\pm0.0126$
      & $30.09\pm0.18$ & $5.310\pm0.007$ & $0.9242\pm0.0119$ \\
    3 & $30.65\pm0.32$ & $5.271\pm0.006$ & $0.9358\pm0.0106$
      & $24.33\pm0.21$ & $5.243\pm0.008$ & $0.9016\pm0.0160$ \\
    4 & $29.30\pm0.24$ & $5.275\pm0.005$ & $0.9176\pm0.0197$
      & \cellcolor[RGB]{232,232,255}$\mathbf{23.56\pm0.30}$
      & \cellcolor[RGB]{232,232,255}$5.241\pm0.007$
      & \cellcolor[RGB]{232,232,255}$0.8896\pm0.0222$ \\
    5 & $29.43\pm0.22$ & $5.268\pm0.006$ & $0.9362\pm0.0063$
      & $25.26\pm0.10$ & $5.247\pm0.004$ & $0.9028\pm0.0075$ \\
    \bottomrule
  \end{tabular}
\end{table}

\subsubsection{XSum Inference-Protocol Selection}
\label{app:xsum-inference-sweep}

We select the classifier-free guidance scale and ODE step count for the XSum comparison using the 300k AURORA-LM-S checkpoint. 
The guidance sweep fixes the sampler at 32 ODE steps, while the step-count sweep fixes the guidance scale at $w=2.0$. 
Each setting is evaluated with five seeds (1--5). 
Table~\ref{tab:app-xsum-inference-sweep} reports the mean and standard deviation across seeds for both sweeps.
Here $w=1.0$ uses the conditional prediction without additional classifier-free guidance.

Moderate guidance improves all three ROUGE metrics over $w=1.0$. 
A scale of $w=2.0$ gives the highest mean ROUGE-1 and ROUGE-L, while $w=1.5$ gives a marginally higher ROUGE-2.
Increasing the scale beyond this range degrades all three metrics. 
With $w=2.0$, 32 ODE steps give the highest mean ROUGE-1 and ROUGE-L, while matching 64 steps in ROUGE-2 after rounding.
Additional steps do not improve the overall ROUGE scores. 
We therefore use 32-step ODE sampling with $w=2.0$ for the XSum comparison.

\begin{table}[htbp]
  \centering
  \footnotesize
  \renewcommand{\arraystretch}{1.05}
  \caption{\textbf{XSum inference-protocol selection for AURORA-LM-S.} Both sweeps
  report mean $\pm$ standard deviation over five sampling seeds. Shaded rows
  denote the selected configuration.}
  \label{tab:app-xsum-inference-sweep}

  \begin{subtable}[t]{0.49\linewidth}
    \centering
    \caption{Guidance scale at 32 ODE steps}
    \scriptsize
    \setlength{\tabcolsep}{2.2pt}
    \begin{tabular}{@{}cccc@{}}
      \toprule
      \textbf{$w$} & \textbf{ROUGE-1 $\uparrow$} & \textbf{ROUGE-2 $\uparrow$} & \textbf{ROUGE-L $\uparrow$} \\
      \midrule
      1.0 & $34.96{\pm}0.34$ & $12.28{\pm}0.13$ & $26.76{\pm}0.30$ \\
      1.5 & $36.42{\pm}0.20$ & $\mathbf{13.47{\pm}0.24}$ & $28.54{\pm}0.25$ \\
      \rowcolor[RGB]{232,232,255}
      2.0 & $\mathbf{36.56{\pm}0.15}$ & $13.43{\pm}0.29$ & $\mathbf{28.86{\pm}0.18}$ \\
      2.5 & $35.87{\pm}0.25$ & $12.78{\pm}0.26$ & $28.37{\pm}0.21$ \\
      3.0 & $34.16{\pm}0.19$ & $11.56{\pm}0.15$ & $26.90{\pm}0.12$ \\
      \bottomrule
    \end{tabular}
  \end{subtable}
  \hfill
  \begin{subtable}[t]{0.49\linewidth}
    \centering
    \caption{ODE steps at $w=2.0$}
    \scriptsize
    \setlength{\tabcolsep}{2.2pt}
    \begin{tabular}{@{}cccc@{}}
      \toprule
      \textbf{Steps} & \textbf{ROUGE-1 $\uparrow$} & \textbf{ROUGE-2 $\uparrow$} & \textbf{ROUGE-L $\uparrow$} \\
      \midrule
      16  & $36.37{\pm}0.22$ & $13.15{\pm}0.26$ & $28.58{\pm}0.24$ \\
      \rowcolor[RGB]{232,232,255}
      32  & $\mathbf{36.56{\pm}0.15}$ & $\mathbf{13.43{\pm}0.29}$ & $\mathbf{28.86{\pm}0.18}$ \\
      64  & $36.45{\pm}0.14$ & $\mathbf{13.43{\pm}0.26}$ & $28.75{\pm}0.14$ \\
      128 & $36.32{\pm}0.17$ & $13.31{\pm}0.26$ & $28.59{\pm}0.19$ \\
      \bottomrule
    \end{tabular}
  \end{subtable}
\end{table}

\FloatBarrier

\subsubsection{Reproduction Protocol for Baselines}

For the OpenWebText unconditional-generation comparison (Sec.~\ref{sec:scale-uncond}), we evaluate released checkpoints from the Duo~\citep{sahoo2025duo} and ELF~\citep{hu2026elf} repositories.
The ELF MAUVE evaluation uses a separate protocol for reference construction and length alignment (Appendix~\ref{app:elf-mauve-protocol}).
For the XSum conditional-generation comparison (Sec.~\ref{sec:scale-cond}), the reported metrics are taken directly from ELF~\citep{hu2026elf}. 
We do not retrain any baseline end to end; results obtained under our evaluation harness differ from the original publications only in the evaluation harness (sampling code, scorer, and metric
implementation), not in the underlying model weights or training recipe.

\begin{table}[!htbp]
\centering
\caption{\textbf{Model architecture and evaluation configurations across baselines.}
The 105M parameters reported for ELF-B in Table~\ref{tab:scale-research} include its backbone and LM head. These exclude the frozen T5-small input
encoder (35M), which is used only during training and not at inference.}
\label{tab:model_arch}
\small
\resizebox{\linewidth}{!}{%
\begin{tabular}{lcccccc}
\toprule
 & \textbf{AR} & \textbf{MDLM} & \textbf{SEDD} & \textbf{DUO} & \textbf{DUO-distilled} & \textbf{ELF-B} \\
\midrule
Hidden size
    & 768    & 768    & 768    & 768    & 768    & 768 \\
Num layers
    & 12     & 12     & 12     & 12     & 12     & 12 \\
Num heads
    & 12     & 12     & 12     & 12     & 12     & 12 \\
Head dim
    & 64     & 64     & 64     & 64     & 64     & 64 \\
Intermediate size
    & 3072   & 3072   & 3072   & 3072   & 3072   & 3072 \\
Vocab size
    & 50257  & 50257  & 50257  & 50257  & 50257  & 32128 \\
Tokenizer
    & GPT-2 BPE & GPT-2 BPE & GPT-2 BPE & GPT-2 BPE & GPT-2 BPE & T5 (SentencePiece) \\
Sequence length
    & 1024   & 1024   & 1024   & 1024   & 1024   & 1024 \\
Sampling steps
    & 1024 tokens & 1024   & 1024   & 1024   & 32     & 32 \\
Sampling method
    & AR (Gumbel-max) & Ancestral & Analytic & Ancestral & Ancestral & SDE \\
\midrule
Backbone params
    & 85.00M & 92.13M & 92.13M & 92.13M & 92.13M & 88M \\
LM head params
    & 38.65M & 38.85M & 38.85M & 38.85M & 38.85M & 17M \\
Vocab embedding params
    & 38.60M & 38.60M & 38.60M & 38.60M & 38.60M & -- \\
\textbf{Total params}
    & \textbf{162.25M} & \textbf{169.58M} & \textbf{169.58M} & \textbf{169.58M} & \textbf{169.58M} & \textbf{105M} \\
\bottomrule
\end{tabular}%
}
\end{table}

AURORA-LM-S uses the same hidden size (768) and depth (12 layers) as the system-level baselines, but its block-causal denoiser contains 130M parameters, compared with approximately 92M for the discrete-diffusion backbones.
This difference primarily reflects its SwiGLU feed-forward layers, which retain the 3,072-dimensional intermediate width while adding
a gating projection, together with the timestep-conditioning pathway and the per-layer self-conditioning projections. 
These backbone counts exclude token embeddings, vocabulary heads, and AURORA-LM's frozen autoencoder.

\subsubsection{Scoring and Answer Extraction}
\label{app:evaluation-protocol}

For OpenWebText free generation, each configuration generates 1,000 samples per
seed over the five evaluation seeds (1--5).
Gen-PPL is computed with the same fixed GPT-2 Large external scorer used in the main text.
The MAUVE scores in Table~\ref{tab:scale-research} use held-out OpenWebText references constructed with GPT-2 wrap preprocessing. We evaluate 1,000 generated and 1,000 reference texts per seed, truncating both to at most 1,024 GPT-2 tokens before extracting GPT-2 Large features. Scores are averaged over five generation seeds.

\noindent\textbf{ELF-B MAUVE reevaluation.}
\phantomsection\label{app:elf-mauve-protocol}
We sample references from the OpenWebText-T5 dataset released with ELF and decode them with T5 to match ELF-B's generated text. ELF-B's generations and these references differ in length, which can affect MAUVE~\citep{shabalin2024tencdm}. To control this difference, we exclude texts shorter than 768 GPT-2 tokens and score the first 768 tokens of 1,000 generations and 1,000 references per seed. Across five seeds, this yields MAUVE $0.768\pm0.062$. We report the result separately because the reference data and truncation differ from those used for Table~\ref{tab:scale-research}; the scores are not directly comparable.

For XSum conditional generation, each model receives the source document as the conditioning prefix and the generated summary is scored with ROUGE. 
For the scaled benchmark profile, AURORA-LM and the released
latent-diffusion reference are evaluated using the same prompt construction,
answer extraction, and metric implementation. Detailed task-specific
conventions are provided in Sec.~\ref{app:benchmark-protocol}.

\subsubsection{Few-Shot Benchmark Protocol}
\label{app:benchmark-protocol}

\noindent\textbf{MMLU.} A multiple-choice benchmark covering humanities, STEM, social sciences, and professional domains~\citep{hendrycks2021mmlu}, used to assess broad factual knowledge and reasoning.

\noindent\textbf{HellaSwag.} A multiple-choice benchmark for grounded commonsense reasoning, requiring selection of the most plausible sentence continuation from adversarially constructed candidates~\citep{zellers2019hellaswag}.

\noindent\textbf{ARC-Challenge.} A multiple-choice benchmark of grade-school science questions selected to be difficult for retrieval- and word-co-occurrence-based baselines, requiring scientific knowledge and reasoning~\citep{clark2018arc}.

\noindent\textbf{WinoGrande.} A large-scale, bias-reduced pronoun-resolution benchmark inspired by the Winograd Schema Challenge, used to assess commonsense reasoning~\citep{sakaguchi2020winogrande}.

\noindent\textbf{OpenBookQA.} A multiple-choice benchmark combining core science facts with broader commonsense knowledge, often requiring multi-hop reasoning~\citep{mihaylov2018openbookqa}.

\noindent\textbf{SIQA.} A social commonsense reasoning benchmark requiring selection of the most plausible response to social situations and intentions~\citep{sap2019siqa}.

\noindent\textbf{StoryCloze.} A story understanding benchmark requiring selection of the most plausible ending for a four-sentence story context, evaluating narrative coherence and causal reasoning~\citep{mostafazadeh2016storycloze}.

\noindent\textbf{RACE.} A reading comprehension benchmark from English examinations, requiring passage understanding and inference beyond simple span extraction~\citep{lai2017race}.

\noindent\textbf{SQuAD.} A reading-comprehension benchmark in which answers are spans in a given passage~\citep{rajpurkar2016squad}. In our evaluation, the model generates a short answer conditioned on the passage and question; we report normalized exact match.

For each benchmark task, we use the standard public evaluation split: the official test split when available and the official validation split otherwise. We construct a fixed candidate subset by shuffling the corresponding split with the default subset-selection seed of 42 and retaining up to 1,000 examples. To ensure matched evaluation, we retain only candidates that satisfy the context-budget requirements of both systems under the shared prompting and generation setup. The resulting fixed task sets are then used unchanged to evaluate both AURORA-LM-L and Cola-DLM. 

We use a common prompt-construction, output-truncation, and answer-matching procedure throughout the benchmark evaluation. Tasks that support in-context demonstrations use two-shot prompts. In multiple-choice demonstrations, answers are written as the full option text
rather than the option letter, following the generative evaluation protocol
used by Cola-DLM~\citep{guo2026cola}.
For multiple-choice tasks, the generated response is truncated at the first
newline, normalized, and mapped to a candidate option; the shared answer
matcher accepts either an explicit option letter or the corresponding option
text. For SQuAD, we apply the same first-newline truncation and standard SQuAD
normalization to both the generated answer and each reference alias:
lowercasing, punctuation and article removal, and whitespace normalization.
We report exact match against the best-matching reference alias.
The exact demonstrations used in these prompts are listed in Appendix~\ref{app:fewshot-demonstrations} (Table~\ref{tab:fewshot-examples}).

\begin{center}
  \footnotesize
  \setlength{\tabcolsep}{4pt}
  \captionof{table}{\textbf{Benchmark evaluation protocol.} All tasks use two-shot
  prompts with textual answer completions. For multiple-choice tasks, the
  scoring parser accepts either the option text or an explicit option letter.}
  \label{tab:benchmark-protocol}
  \begin{tabular}{lccc}
    \toprule
    \textbf{Task} & \textbf{Shots} & \textbf{Accepted output} &
    \textbf{Scoring} \\
    \midrule
    HellaSwag & 2 & Option text (or letter) & Accuracy \\
    MMLU & 2 & Option text (or letter) & Accuracy \\
    OpenBookQA & 2 & Option text (or letter) & Accuracy \\
    RACE & 2 & Option text (or letter) & Accuracy \\
    SIQA & 2 & Option text (or letter) & Accuracy \\
    SQuAD & 2 & Short answer text & Normalized exact match (EM) \\
    StoryCloze & 2 & Option text (or letter) & Accuracy \\
    ARC-Challenge & 2 & Option text (or letter) & Accuracy \\
    WinoGrande & 2 & Option text (or letter) & Accuracy \\
    \bottomrule
  \end{tabular}
\end{center}

\FloatBarrier
\subsection{Additional Numerical Results for Controlled Design Analysis}
\label{app:design-study-diagnostics}

This section reports the exact numerical results underlying the controlled OWT128 analyses in Sec.~\ref{sec:study}.
Unless otherwise stated, evaluations use five generation seeds (1--5), with 1,000 samples generated in each evaluation run.
Tables report the mean $\pm$ standard deviation across seeds, and generation quality is measured using MAUVE against held-out OpenWebText references.
\subsubsection{Latent Capacity and Sequence Compression}
\label{app:full-diagnostics}

\begin{center}
  \begin{minipage}{\linewidth}
  \centering
  \scriptsize
  \setlength{\tabcolsep}{3.2pt}
  \captionof{table}{\textbf{Full channel-width and tan-$d$ schedule matrix.} Each cell reports mean MAUVE $\pm$ standard deviation over evaluation seeds. Bold marks the best schedule for each latent width.}
  \label{tab:app-channel-matrix}
  \vspace{0.3em}
  \begin{tabular}{ccccc}
    \toprule
    \textbf{Latent channel $D$} & \textbf{$d=1$} & \textbf{$d=3$} & \textbf{$d=5$} & \textbf{$d=7$} \\
    \midrule
    128  & $0.345{\pm}0.027$ & $0.578{\pm}0.025$ & $\mathbf{0.594{\pm}0.035}$ & $0.509{\pm}0.039$ \\
    256  & $0.230{\pm}0.012$ & $0.616{\pm}0.025$ & $0.604{\pm}0.031$ & $\mathbf{0.640{\pm}0.033}$ \\
    512  & $0.236{\pm}0.018$ & $0.637{\pm}0.022$ & $\mathbf{0.693{\pm}0.011}$ & $0.667{\pm}0.036$ \\
    1024 & $0.229{\pm}0.023$ & $0.797{\pm}0.032$ & $0.815{\pm}0.012$ & $\mathbf{0.839{\pm}0.024}$ \\
    \bottomrule
  \end{tabular}
  \end{minipage}

  \medskip

  \begin{minipage}{\linewidth}
  \centering
  \scriptsize
  \setlength{\tabcolsep}{6.0pt}
  \captionof{table}{\textbf{Sequence-compression results underlying Fig.~\ref{fig:study-latent-channel}(c).} The latent channel width is fixed at $D=1024$ and the noisy-input bottleneck at $D_b=128$.}
  \label{tab:study-compression}
  \vspace{0.3em}
  \begin{tabular}{cc}
  \toprule
  \textbf{Latent positions retained} & \textbf{MAUVE $\uparrow$} \\
  \midrule
  $100\%$ & $0.815\pm0.012$ \\
  $90\%$  & $0.807\pm0.022$ \\
  $80\%$  & $0.726\pm0.020$ \\
  $70\%$  & $0.671\pm0.063$ \\
  \bottomrule

  \end{tabular}
  \end{minipage}
\end{center}
\FloatBarrier

\subsubsection{Modeling the Full-Width Latent Distribution}
\label{app:full-width-diagnostics}

\begin{center}
  \begin{minipage}{\linewidth}
  \centering
  \scriptsize
  \setlength{\tabcolsep}{3.2pt}
  \captionof{table}{\textbf{Detailed noisy-input bottleneck sweep for a full-width clean target with $D=1024$.} Each step-specific entry reports mean MAUVE $\pm$ standard deviation over evaluation seeds. The final column averages the three step-specific means and is the statistic shown in Fig.~\ref{fig:study-bottleneck-schedule}(a).}
  \label{tab:study-bottleneck}
  \vspace{0.3em}
  \begin{tabular}{ccccc}
  \toprule
  \textbf{$D_b$} & \textbf{16 steps} & \textbf{32 steps}
  & \textbf{64 steps} & \textbf{Mean} \\
  \midrule
  32   & $0.793{\pm}0.019$ & $0.795{\pm}0.009$ & $0.771{\pm}0.025$ & 0.786 \\
  128  & $0.774{\pm}0.013$ & $0.815{\pm}0.012$ & $0.835{\pm}0.027$ & \textbf{0.808} \\
  256  & $0.728{\pm}0.025$ & $0.797{\pm}0.046$ & $0.754{\pm}0.023$ & 0.760 \\
  512  & $0.706{\pm}0.043$ & $0.779{\pm}0.023$ & $0.759{\pm}0.047$ & 0.748 \\
  1024 & $0.757{\pm}0.017$ & $0.807{\pm}0.033$ & $0.807{\pm}0.014$ & 0.790 \\
  \bottomrule
\end{tabular}
  \end{minipage}
\end{center}
\FloatBarrier

\begin{figure*}[h]
    \centering
  \includegraphics[width=0.9\linewidth]{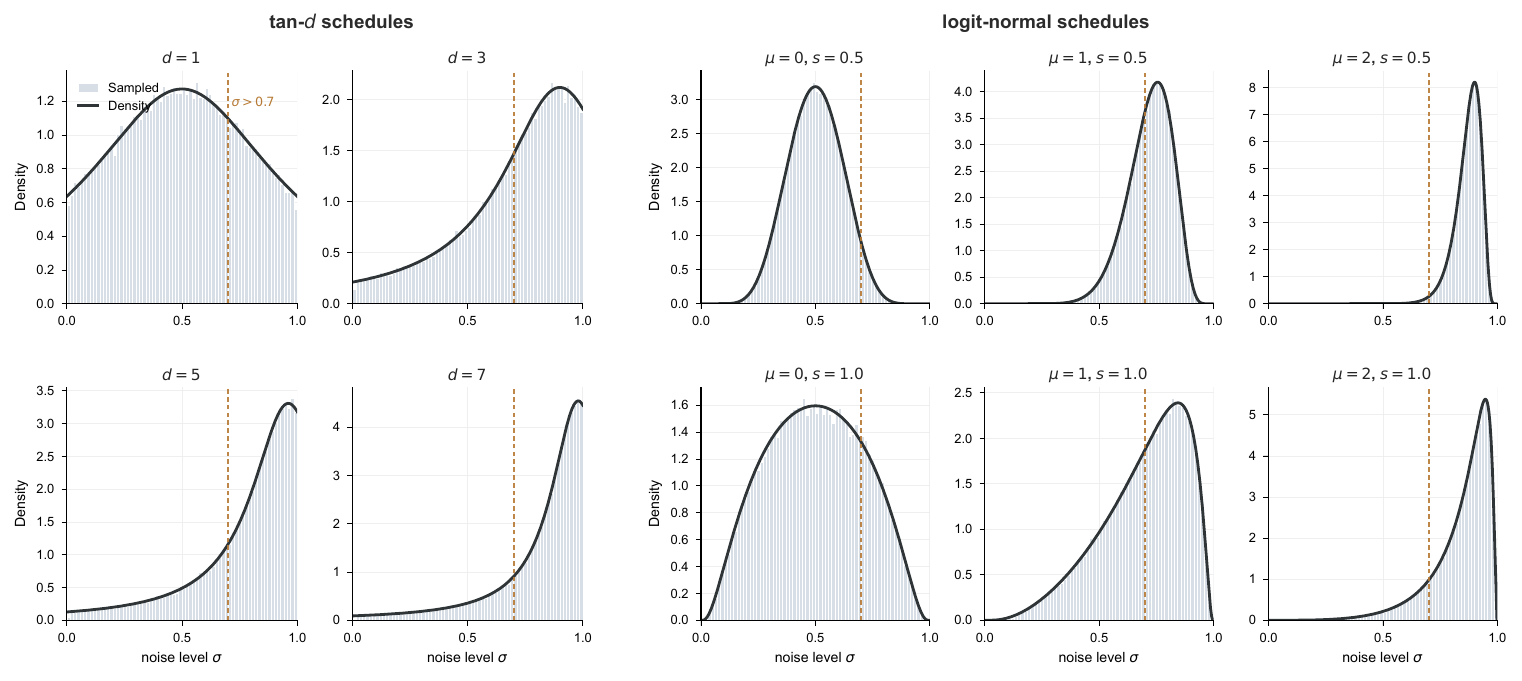}
  \captionof{figure}{\textbf{Training-time noise distributions used in the schedule ablation.}
  Each panel shows the density over the flow-matching noise level $\sigma\in[0,1]$.
  The dashed reference marks the high-noise region $\sigma>0.7$ used in
  Fig.~\ref{fig:study-bottleneck-schedule}(b).}
  \label{fig:app-schedule-distributions}
\end{figure*}

\phantomsection
\label{app:schedule-distributions}
\noindent\textbf{Noise allocation.}
Figure~\ref{fig:app-schedule-distributions} visualizes the training-time noise distributions used in the schedule ablation. 
Table~\ref{tab:app-schedule-allocation} reports the corresponding high-noise mass and generation quality.

\FloatBarrier

\begin{center}
  \begin{minipage}{\linewidth}
  \centering
  \scriptsize
  \setlength{\tabcolsep}{5.0pt}
  \captionof{table}{\textbf{Noise-allocation sweep underlying Fig.~\ref{fig:study-bottleneck-schedule}(b).} The high-noise mass is $m_{0.7}=\Pr(\sigma>0.7)$; MAUVE is reported as mean $\pm$ standard deviation over evaluation seeds.}
  \label{tab:app-schedule-allocation}
  \vspace{0.3em}
  \begin{tabular}{lccc}
    \toprule
    \textbf{Schedule family} & \textbf{Parameter} & \textbf{$m_{0.7}$} & \textbf{MAUVE $\uparrow$} \\
    \midrule
    tan-$d$ & $d=1$ & 0.258 & $0.229{\pm}0.023$ \\
    tan-$d$ & $d=3$ & 0.579 & $0.797{\pm}0.032$ \\
    tan-$d$ & $d=5$ & 0.722 & $0.815{\pm}0.012$ \\
    tan-$d$ & $d=7$ & 0.795 & $\mathbf{0.839{\pm}0.024}$ \\
    \midrule
    logit-normal, $s=0.5$ & $\mu=0$ & 0.045 & $0.010{\pm}0.001$ \\
    logit-normal, $s=0.5$ & $\mu=1$ & 0.620 & $0.111{\pm}0.011$ \\
    logit-normal, $s=0.5$ & $\mu=2$ & 0.989 & $0.828{\pm}0.017$ \\
    \midrule
    logit-normal, $s=1.0$ & $\mu=0$ & 0.198 & $0.022{\pm}0.003$ \\
    logit-normal, $s=1.0$ & $\mu=1$ & 0.561 & $0.586{\pm}0.051$ \\
    logit-normal, $s=1.0$ & $\mu=2$ & 0.875 & $0.833{\pm}0.014$ \\
    \bottomrule
  \end{tabular}
  \end{minipage}
\end{center}
\FloatBarrier

\subsubsection{Prediction Target and Loss Space}
\label{app:target-loss-diagnostics}

\begin{center}
  \begin{minipage}{\linewidth}
  \centering
  \scriptsize
  \setlength{\tabcolsep}{5.0pt}
  \captionof{table}{\textbf{Target and loss-space analysis for a full-width clean target with $D=1024$.} Direct clean-latent regression yields the highest MAUVE under both noisy-input widths.}
  \label{tab:app-target-loss}
  \vspace{0.3em}
  \begin{tabular}{ccrr}
    \toprule
    & & \multicolumn{2}{c}{\textbf{Noisy-input width}} \\
    \cmidrule(lr){3-4}
    \textbf{Target} & \textbf{Loss} & $D_b=128$ & $D_b=1024$ \\
    \midrule
    \multirow{2}{*}{$x_0$}
      & $x_0$ & $\mathbf{0.815\pm0.012}$ & $\mathbf{0.807\pm0.033}$ \\
      & $v$   & $0.059\pm0.009$ & $0.017\pm0.002$ \\
    \multirow{2}{*}{$v$}
      & $x_0$ & $0.729\pm0.023$ & $0.325\pm0.021$ \\
      & $v$   & $0.653\pm0.034$ & $0.301\pm0.016$ \\
    \bottomrule
  \end{tabular}
  \end{minipage}
\end{center}
\FloatBarrier

\subsubsection{Few-Step and Blockwise Generation}
\label{app:blockwise-generation-diagnostics}

\begin{table}[htbp]
  \centering
  \caption{\textbf{Exact numerical results underlying Fig.~\ref{fig:study-trajectory-block}.} Values are mean MAUVE $\pm$ standard deviation over evaluation seeds.}
  \label{tab:app-study-inference}
  \footnotesize
  \renewcommand{\arraystretch}{1.05}
  \setlength{\tabcolsep}{5.0pt}

  \begin{subtable}[t]{0.54\linewidth}
    \centering
    \caption{Self-trajectory consistency across denoising-step budgets}
    \begin{tabular}{@{}ccc@{}}
      \toprule
      \textbf{Steps} & \textbf{Consistency off} & \textbf{Consistency on} \\
      \midrule
      8   & $0.116{\pm}0.006$ & $0.672{\pm}0.038$ \\
      16  & $0.310{\pm}0.039$ & $0.786{\pm}0.034$ \\
      32  & $0.589{\pm}0.062$ & $0.815{\pm}0.012$ \\
      64  & $0.660{\pm}0.023$ & $\mathbf{0.832{\pm}0.014}$ \\
      128 & $\mathbf{0.757{\pm}0.029}$ & $0.825{\pm}0.024$ \\
      256 & $0.724{\pm}0.022$ & $0.797{\pm}0.014$ \\
      \bottomrule
    \end{tabular}
  \end{subtable}
  \hfill
  \begin{subtable}[t]{0.40\linewidth}
    \centering
    \caption{Trained block size at 32 ODE steps}
    \begin{tabular}{@{}cc@{}}
      \toprule
      \textbf{$Q$} & \textbf{MAUVE $\uparrow$} \\
      \midrule
      4  & $\mathbf{0.909{\pm}0.012}$ \\
      8  & $0.857{\pm}0.011$ \\
      16 & $0.815{\pm}0.012$ \\
      32 & $0.681{\pm}0.023$ \\
      64 & $0.631{\pm}0.027$ \\
      \bottomrule
    \end{tabular}
  \end{subtable}
\end{table}

\FloatBarrier


\subsection{Qualitative OWT Examples}
\label{app:owt-examples}

To give a concrete view of AURORA-LM's unconditional generation behavior, we include two examples from the OWT1024 free-generation evaluation. 
The Gen-PPL and Entropy values shown alongside each example are computed by the fixed GPT-2 Large external scorer for that
individual sample; the main quantitative comparison is based on the corpus-level results in
Table~\ref{tab:scale-research}.

\medskip
\noindent
\fcolorbox{auroraPurple}{auroraLavender}{%
\begin{minipage}{0.94\linewidth}
\small
\textcolor{auroraPurple}{\textbf{Example 1}}\hfill
\small PPL: 24.625 \quad Entropy: 5.467
\par\smallskip
{\normalsize
\textbf{Generation.}\par\smallskip
\ttfamily
<|endoftext|> When talking about open source, there's a few options at play: open access to an existing project, give to existing members of the community a chance to continue what they've loved, or use it to develop code and features used by others.\par\medskip

Of course, there have been a litany of different options: for instance, in the past, Ruby contributors could just write Ruby code, then publish it, on the forums, include the source code, then use it for others to improve their code if they like. Well, this sounds logical to me. It would open up the community at the opportunity new people could use and improve on their original code without having to go through all of the large hoops that come with trying to provide good quality code.\par\medskip

Unfortunately, open source seems to be a possibility, so here comes Direct Air, a project run by a guy named Justin Vipme. He's the most devoted author of what looks like an entirely free, wide-used code base, without any permission.\par\medskip

Two years ago, Justin sent me an email I wanted to give you at least a overview of it. The first part I think here brought me an immediate interest in some module creation. Yes, you see, module creation---also called "backCreating"---is the complex process that takes place at any time in your application. Basically, working between the users and the developers is separate from working on your own system. On the other hand, the first thing I wish I could think of doing with Direct Air would be doing a bug report to the bug delivery team and then release a solution to make sure the project works right from the ground up.\par\medskip

So that's what surprised me at the time. You'll need to speak to some open source author running anything other than Ruby in order to make a bug report:\par\smallskip
https://github.com/merick-vipme/backCreating.im\par\medskip

This involved taking a tip of each module that went wrong and trying to pin down how they were fixed. In many ways, the project actually helped make the process run faster and more efficient than many people had hoped. And it also made it much easier to developers to issue a bug report and make sure the module works as intended. Plus, a combination of each module that allows them to automatically communicate with the developers involved made it easy for them to report an issue in their own system more readily and efficiently, rather than relying on a message engine.\par\medskip

You can see this happening in the make-up of the bug report below, and when you do, you should get a commit line that shows just which of the modules that went wrong was fixed.\par\medskip

As you can see, Vipme was dealing with 75\% of bugs, which makes it easier and faster for developers to check if they're worth patches. And to keep in mind, this doesn't mean that somebody has tried to open a module or that somebody has accidentally written your package. It does mean that reaching out to others will help ensure that bugs aren't being fixed very easily.\par\medskip

The coolest part is that the bug reports\par\smallskip
<|endoftext|>\par\smallskip}
\end{minipage}}

\clearpage
\begin{tcolorbox}[
  breakable,
  colback=auroraLavender,
  colframe=auroraPurple,
  boxrule=0.5pt,
  left=6pt,
  right=6pt,
  top=6pt,
  bottom=6pt
]
\small
\textcolor{auroraPurple}{\textbf{Example 2}}\hfill
\small PPL: 22.00 \quad Entropy: 5.358
\par\smallskip
{\normalsize
\textbf{Generation.}\par\smallskip
\ttfamily
<|endoftext|> mediators have to carve out a compromise that is less likely than most. They will agree on strong demands as early as Monday or early Friday unless external demands are met, government officials said.\par\medskip

They will also agree on rights to limiting refugee arrivals in Belgium and stopping people from trying to travel across the Atlantic to countries including Sweden and France.\par\medskip

The EU is close to setting up talks with Russia, paving the way for similar talks with the UK, France and Italy in July and September of a need for more sharing. Those talks have already started in Geneva this week and in London.\par\medskip

George Murphy, the head of the European Commission, said that Britain should be "perceived" with an agreement on immigration, security and criminal justice on Wednesday.\par\medskip

But he said the crisis had not developed overnight. Europe should also look at the case-by-case role of external mediators in securing a solution.\par\medskip

Speaking from the EU headquarters, he acknowledged that the global migrant crisis is complex and said it was too early to resolve it.\par\medskip

He said the UK should not waste the opportunity of persuading all EU nations to introduce new measures, such as cutting access to shelters, in order to prevent Islamic State fighters from crossing the border.\par\medskip

He also suggested it should set up central migrant crisis directors' office in Belgium with experts on security, criminal justice and democracy as well as human rights.\par\medskip

EU officials indicated Thursday that representatives from would have to co-operate with 12 member states in order to come up with a strategy to meet their demands.\par\medskip

The number of asylum claims has more than doubled in the past two years and doubled in 2016, with more than 6,450 people now entering Europe after a decade of obstacles. There have been complaints about ticket queues at the border too long and locals say tourists are being denied housing.\par\medskip

David Cameron: 'We need to make sure not a single migrant comes in safely'\par\medskip

There has been an unprecedented surge of refugees coming to make it to Europe over the past year. There are between 9,000 and 5,000 a day a day from Iraq, Syria, Yemen, Yemen and Somalia. More than 5 million people live on the EU's Southern border which provides a major hub for smuggling and support Islamic State fighters in Syria and Somalia - from Iraq and Syria.\par\medskip

There have also been an unprecedented number of refugees arriving from Turkey over the past year which are about 268 refugees a day, what Cameron termed "the so-called country capital".\par\medskip

Mr Cameron, whose remarks were at Downing Street on Monday, said Belgium should always "co-operate with other countries to try to resolve the migrant crisis". He also repeated the call for external mediators to be involved in figuring out ways to meet the social and economic challenges that are roiling the West.\par\medskip

He told Sky News: "So I'm not being framed because that sounds like George Murphy. I think what this Chancellor wants to do is to try to carve out a complex structure with other countries to try to resolve things that are less likely than most."\par\medskip

In his first briefing on refugees a EU spokeswoman said: "There have been a wave of people attempting to leave Syria or Libya, join Syria, wanted to cross the border to join IS or other Islamic extremist groups or expressed a desire or willing to turn over to join terrorist groups carrying out attacks on civilians.\par\medskip

"Our aim is to broadenen the negotiating principles with all nations to come up with a common structure so that other countries can work together to resolve the crisis across the globe.\par\medskip

"The number of people arriving into Europe has risen more than doubled in the past four years. It is the responsibility of each member to support the handling of the global migrant crisis.\par\medskip

World of freedom\par\medskip

After Monday's cabinet meeting, Belgium was still grappling to come to terms with a deal that would allow a refugee to cross the EU within two months.\par\medskip

But in his first statement, the chancellor said: "I assure you that this is the world of freedom: where no matter of ethnicity, race, religion, national origin or national belief we can turn the crisis from one country to another and work together to work together to find a common solution."\par\medskip

EU refugee and migrant minister Jean-Claud Juncker also said cross-border talks between the 58 members needed to take place before they could come up with a common structure to resolve the crisis.\par\medskip

Belgian immigration minister Step-Luc Lassier warned the long-term effects of the crisis could be "less likely than most" for refugees fleeing the West and elsewhere across the globe.\par\medskip

More than 5,000 foreigners attempt to cross the EU every day from Syria and Iraq - many of them come from Afghanistan and Pakistan, which accounts for nearly a third of the EU's total flights, according to the UN Office for Refugees.\par\medskip

UNICE figures show there are 450,000 foreigners attempting to cross the EU from Syria and Iraq every day.\par\medskip

<|endoftext|>\par\smallskip}
\end{tcolorbox}

\FloatBarrier

\subsection{Qualitative XSum Examples}
\label{app:xsum-examples}

To give a concrete view of AURORA-LM's behavior on XSum, we include two examples from the evaluation set. 
Each example pairs the complete input article with its reference summary and the summary produced by our model. 
The ROUGE-F1 values shown alongside the examples are computed for the individual pairs; the main quantitative comparison is based on the corpus-level results in Table~\ref{tab:scale-xsum}.

\noindent
\fcolorbox{auroraPurple}{auroraLavender}{%
\begin{minipage}{0.94\linewidth}
\small
\textcolor{auroraPurple}{\textbf{Example 1}}\hfill
\small ROUGE-F1 (R-1 / R-2 / R-L): 60.5 / 34.1 / 55.8
\par\smallskip
{\footnotesize
\textbf{Source.} After a drab game with few chances for either side, Mohamed
Salah teed up El Said, who struck firmly past Uganda keeper Denis Onyango on 89
minutes. Uganda's Joseph Ochaya earlier had a goal correctly disallowed for
offside. The win means Egypt play Ghana on Wednesday with top spot of Group D
still at stake, while Uganda are out of their first tournament since 1978. Ghana
secured their place in the quarter-finals with a 1--0 victory over Mali earlier
on Saturday. On a deteriorating pitch in Port-Gentil, seven-times champions
Egypt failed to break down a resolute Cranes defence for long periods. Onyango
made an impressive clearing header to deny the onrushing Salah at the end of an
otherwise lifeless first half. Uganda began the second half brightly, forcing
veteran Egypt goalkeeper Essam El Hadary to punch several set-pieces clear but
ultimately failed to register a shot on target as Ochaya turned in Faruku Miya's
through ball from a clearly offside position on 52 minutes. El Said blasted a
shot well over when unmarked and Marwan Mohsen went close for the Pharaohs with
a header, but Egypt appeared to be heading for their second consecutive
goalless draw. However, an otherwise quiet Salah showed great composure late on
to check his shot and play in El Said with a reverse pass, the substitute's
close-range shot evading Onyango. Uganda now face Mali in the final round of
group games on Wednesday, with Mali needing a win and for Ghana to beat Egypt by
two goals or more if they are to qualify. Match ends, Egypt 1, Uganda 0. Second
Half ends, Egypt 1, Uganda 0. Offside, Uganda. Murshid Jjuko tries a through
ball, but Muhammad Shaban is caught offside. Hand ball by Moses Oloya (Uganda).
Kahraba (Egypt) wins a free kick in the attacking half. Foul by Hassan Wasswa
(Uganda). Offside, Egypt. Kahraba tries a through ball, but Marwan Mohsen is
caught offside. Substitution, Uganda. Moses Oloya replaces Denis Iguma. Mohamed
Shafy (Egypt) wins a free kick in the defensive half. Foul by Khalid Aucho
(Uganda). Goal! Egypt 1, Uganda 0. Abdallah El Said (Egypt) right footed shot
from the centre of the box to the centre of the goal. Assisted by Mohamed Salah.
Attempt missed. Amr Warda (Egypt) right footed shot from outside the box is too
high. Assisted by Mohamed Shafy. Corner, Uganda. Conceded by Ahmed Fathy.
Attempt blocked. Joseph Ochaya (Uganda) left footed shot from outside the box is
blocked. Assisted by Nicholas Wadada. Delay over. They are ready to continue.
Substitution, Uganda. Nicholas Wadada replaces Faruku Miya. Substitution, Egypt.
Kahraba replaces Trezeguet. Delay in match Khalid Aucho (Uganda) because of an
injury. Attempt saved. Marwan Mohsen (Egypt) right footed shot from outside the
box is saved in the centre of the goal. Assisted by Amr Warda. Tonny Mawejje
(Uganda) wins a free kick in the defensive half. Foul by Mohamed Elneny
(Egypt). Delay over. They are ready to continue. Delay in match Mohamed Shafy
(Egypt) because of an injury. Foul by Muhammad Shaban (Uganda). Ali Gabr
(Egypt) wins a free kick in the defensive half. Hand ball by Trezeguet (Egypt).
Attempt missed. Marwan Mohsen (Egypt) header from the centre of the box is just
a bit too high. Assisted by Mohamed Shafy with a cross following a set piece
situation. Foul by Murshid Jjuko (Uganda). Marwan Mohsen (Egypt) wins a free
kick in the attacking half. Godfrey Walusimbi (Uganda) wins a free kick on the
left wing. Foul by Ahmed Fathy (Egypt). Attempt saved. Trezeguet (Egypt) right
footed shot from outside the box is saved in the centre of the goal.
Attempt blocked.\par\smallskip}
\normalsize\textbf{Generation.} Egypt's Abdallah El Said scored an injury-time
winner to beat hosts Uganda and reach the Africa Cup of Nations
quarter-finals.\par\smallskip
\normalsize\textbf{Reference.} Substitute Abdallah El Said scored a late winner
for Egypt to knock Uganda out of the Africa Cup of Nations.
\end{minipage}}

\medskip
\noindent
\fcolorbox{auroraPurple}{auroraLavender}{%
\begin{minipage}{0.94\linewidth}
\small
\textcolor{auroraPurple}{\textbf{Example 2}}\hfill
\small ROUGE-F1 (R-1 / R-2 / R-L): 53.3 / 35.7 / 53.3
\par\smallskip
{\normalsize
\textbf{Source.} The latest picture was taken by the rover at ``Buckskin,''
which is the seventh rock target on its mission. Curiosity does this planetary
photography in the same way we would take a selfie - by holding a camera at
arm's length and framing itself. The pictures must then be stitched together to
make this final scene. Nasa is trying to take one at every location where the
robot drills into the surface of the red planet. The robot has now been on Mars
for three Earth years. It is in what's known as Gale Crater. Curiosity is
currently climbing through what's known as Mount Sharp, examining the rocks as
it goes. Scientists want to understand when and how Mars became so barren.\par\smallskip}
\normalsize\textbf{Generation.} The US space agency, Nasa, has taken a stunning
picture of the Mars robot.\par\smallskip
\normalsize\textbf{Reference.} The US space agency Nasa has issued a ``selfie''
portrait from its Curiosity rover on Mars.
\end{minipage}}

\FloatBarrier

\subsection{Few-Shot Benchmark Demonstrations}
\label{app:fewshot-demonstrations}

To give a concrete view of our 1B-parameter AURORA-LM's few-shot evaluation protocol, we list the two-shot demonstrations used for each benchmark task in Sec.~\ref{sec:scale-benchmark}. 
The quantitative comparison is reported in Table~\ref{tab:scale-benchmark}.

\begingroup
  \footnotesize
  \setlength{\tabcolsep}{2.5pt}
  \renewcommand{\arraystretch}{0.95}
  \setlength{\fboxsep}{24pt}
  \begin{table}[!htbp]
    \centering
    \caption{Two-shot demonstrations used for each benchmark task.}
    \label{tab:fewshot-examples}
    \fcolorbox{auroraPurple}{auroraLavender}{%
    \begin{minipage}{0.93\linewidth}
    \centering
    \begin{tabular}{p{0.47\linewidth} p{0.47\linewidth}}
      \toprule
      \textbf{Example 1} & \textbf{Example 2} \\
      \midrule
      \multicolumn{2}{l}{\textit{HellaSwag}} \\
      \parbox[t]{0.47\linewidth}{\ttfamily Context: The girl puts the bread into the toaster and pushes the lever down. The bread\\ (A) becomes a slice of pizza.\ (B) starts to toast and turn brown.\ (C) disappears immediately.\ (D) turns into a glass of water.\\ Answer: starts to toast and turn brown.}
      &
      \parbox[t]{0.47\linewidth}{\ttfamily Context: The goalkeeper sees the ball coming towards the net. He dives and\\ (A) catches the ball with his hands.\ (B) starts dancing in the field.\ (C) opens a laptop to check email.\ (D) runs away from the stadium.\\ Answer: catches the ball with his hands.} \\
      \midrule
      \multicolumn{2}{l}{\textit{MMLU}} \\
      \parbox[t]{0.47\linewidth}{\ttfamily Question: Which gas do plants absorb from the air during photosynthesis?\\ (A) Oxygen\ (B) Carbon dioxide\ (C) Nitrogen\ (D) Hydrogen\\ Answer: Carbon dioxide}
      &
      \parbox[t]{0.47\linewidth}{\ttfamily Question: What is the capital of France?\\ (A) London\ (B) Berlin\ (C) Paris\ (D) Madrid\\ Answer: Paris} \\
      \midrule
      \multicolumn{2}{l}{\textit{OpenBookQA}} \\
      \parbox[t]{0.47\linewidth}{\ttfamily Question: Which tool is best for tightening a screw?\\ (A) spoon\ (B) hammer\ (C) screwdriver\ (D) paintbrush\\ Answer: screwdriver}
      &
      \parbox[t]{0.47\linewidth}{\ttfamily Question: What do plants absorb from the air during photosynthesis?\\ (A) carbon dioxide\ (B) oxygen\ (C) helium\ (D) salt\\ Answer: carbon dioxide} \\
      \bottomrule
    \end{tabular}
    \end{minipage}}%
  \end{table}
  \clearpage
  \begin{table}[htbp]
    \centering
    \caption*{Two-shot demonstrations (continued).}
    \fcolorbox{auroraPurple}{auroraLavender}{%
    \begin{minipage}{0.93\linewidth}
    \centering
    \begin{tabular}{p{0.47\linewidth} p{0.47\linewidth}}
      \toprule
      \textbf{Example 1} & \textbf{Example 2} \\
      \midrule
      \multicolumn{2}{l}{\textit{RACE}} \\
      \parbox[t]{0.47\linewidth}{\ttfamily Article: Mary went to the store to buy some fruits. She bought five apples and two oranges. She paid 5 dollars in total.\\ Question: What did Mary buy?\ Options:\ (A) Bananas\ (B) Apples and oranges\ (C) Grapes\ (D) Watermelon\\ Answer: Apples and oranges}
      &
      \parbox[t]{0.47\linewidth}{\ttfamily Article: Sarah loves reading books. She goes to the library every Saturday morning. Last Saturday, she borrowed three mystery novels and two science fiction books.\\ Question: How many books did Sarah borrow last Saturday?\ Options:\ (A) Two\ (B) Three\ (C) Four\ (D) Five\\ Answer: Five} \\
      \midrule
      \multicolumn{2}{l}{\textit{SIQA}} \\
      \parbox[t]{0.47\linewidth}{\ttfamily Question: Jordan wanted to tell a joke to his friends. What does Jordan need to do before this?\\ (A) ignore his friends\ (B) think of a funny story\ (C) leave the room\\ Answer: think of a funny story}
      &
      \parbox[t]{0.47\linewidth}{\ttfamily Question: Tom forgot his umbrella on a rainy day. How would Tom feel?\\ (A) happy\ (B) excited\ (C) frustrated\\ Answer: frustrated} \\
      \bottomrule
    \end{tabular}
    \end{minipage}}%
  \end{table}
  \clearpage
  \begin{table}[htbp]
    \centering
    \caption*{Two-shot demonstrations (continued).}
    \fcolorbox{auroraPurple}{auroraLavender}{%
    \begin{minipage}{0.93\linewidth}
    \centering
    \begin{tabular}{p{0.47\linewidth} p{0.47\linewidth}}
      \toprule
      \textbf{Example 1} & \textbf{Example 2} \\
      \midrule
      \multicolumn{2}{l}{\textit{SQuAD}} \\
      \parbox[t]{0.47\linewidth}{\ttfamily Context: The Normans (Norman: Nourmands; French: Normands; Latin: Normanni) were the people who in the 10th and 11th centuries gave their name to Normandy, a region in France. They were descended from Norse raiders and pirates from Denmark, Iceland and Norway.\\ Question: In what country is Normandy located?\\ Answer: France}
      &
      \parbox[t]{0.47\linewidth}{\ttfamily Context: The Apollo program, also known as Project Apollo, was the third United States human spaceflight program carried out by NASA, which accomplished landing the first humans on the Moon from 1969 to 1972.\\ Question: Which organization carried out the Apollo program?\\ Answer: NASA} \\
      \bottomrule
    \end{tabular}
    \end{minipage}}%
  \end{table}
  \clearpage
  \begin{table}[htbp]
    \centering
    \caption*{Two-shot demonstrations (continued).}
    \fcolorbox{auroraPurple}{auroraLavender}{%
    \begin{minipage}{0.93\linewidth}
    \centering
    \begin{tabular}{p{0.47\linewidth} p{0.47\linewidth}}
      \toprule
      \textbf{Example 1} & \textbf{Example 2} \\
      \midrule
      \multicolumn{2}{l}{\textit{StoryCloze}} \\
      \parbox[t]{0.47\linewidth}{\ttfamily Story: I wanted to make an omelet. I cracked two eggs into a bowl and whisked them. Then I poured them into a hot pan.\\ (A) I ate a delicious omelet for breakfast.\ (B) I decided to order a pizza instead.\\ End: I ate a delicious omelet for breakfast.}
      &
      \parbox[t]{0.47\linewidth}{\ttfamily Story: The runner tied his shoes tight. He sprinted as fast as he could during the race. He crossed the finish line first.\\ (A) He was sad that he lost the race.\ (B) He won the gold medal.\\ End: He won the gold medal.} \\
      \midrule
      \multicolumn{2}{l}{\textit{ARC-Challenge}} \\
            \parbox[t]{0.47\linewidth}{\ttfamily Question: Which of the following best explains why a helium balloon floats in air?\\ (A) Helium is lighter than the rubber of the balloon.\ (B) Helium has a lower density than air.\ (C) Helium atoms are larger than air molecules.\ (D) Helium reacts with the air to produce lift.\\ Answer: Helium has a lower density than air.}
      &
      \parbox[t]{0.47\linewidth}{\ttfamily Question: Which process is responsible for the movement of water from plant roots to leaves?\\ (A) Active transport only\ (B) Gravity\ (C) Transpiration pull\ (D) Osmosis only\\ Answer: Transpiration pull} \\
      \midrule
      \multicolumn{2}{l}{\textit{WinoGrande}} \\
      \parbox[t]{0.47\linewidth}{\ttfamily Sentence: The cup didn't fit in the suitcase because \_ was too big.\ (A) the cup\ (B) the suitcase\\ Answer: the cup}
      &
      \parbox[t]{0.47\linewidth}{\ttfamily Sentence: The trophy didn't fit into the suitcase because \_ was too small.\ (A) the trophy\ (B) the suitcase\\ Answer: the suitcase} \\
      \bottomrule
    \end{tabular}
    \end{minipage}}%
  \end{table}
\endgroup
\FloatBarrier

\end{document}